\documentclass[journal]{IEEEtran}

\ifCLASSINFOpdf
	\usepackage{graphics} 
	\usepackage{epsfig} 
	\usepackage{mathptmx} 
	\usepackage{times} 
	\usepackage{amsmath} 
	\usepackage{amssymb}  
	\usepackage{textcomp}
	\usepackage{multirow}
	\usepackage{colortbl}
	\usepackage{xcolor}
	\usepackage{hhline}
	\usepackage{subfigure}
	\usepackage{bm}
	\usepackage{hyperref}
	\usepackage{lipsum}
	\usepackage{mathtools}
	\usepackage{cuted}
	\usepackage{algorithm}
    	\usepackage{algpseudocode}
	\graphicspath{{figs/}{evaluation/}}
\else
\fi

\renewcommand{\vec}[1]{\mathbf{#1}}
\newcommand\norm[1]{\left\lVert#1\right\rVert}

\usepackage[colorinlistoftodos, textwidth=3.7cm]{todonotes}
\definecolor{MyBrickRed}{cmyk}{0,0.89,0.94,0.28}

\begin{document}

\title{Inverse RL Scene Dynamics Learning for Nonlinear Predictive Control in Autonomous Vehicles}

\author{Sorin~Grigorescu and Mihai~Zaha
\vspace{-1.5em}
\thanks{Manuscript received May 19, 2024; revised 18 September 2024; accepted 03 February 2025. The authors would like to thank Elektrobit Automotive for the infrastructure and research support.}
\thanks{The authors are with the Robotics, Vision and Control Laboratory (RovisLab) at the Department of Automation and Information Technology, Transilvania University of Brasov, 500036 Romania and Elektrobit Automotive. e-mail: (see \url{http://rovislab.com/sorin_grigorescu.html}).}
}

\markboth{IEEE Transactions on Neural Networks and Learning Systems, March 2025}%
{Grigorescu \MakeLowercase{\textit{et al.}}: Inverse RL Scene Dynamics Learning for Nonlinear Predictive Control in Autonomous Vehicles}

\maketitle

\begin{abstract}

This paper introduces the Deep Learning-based Nonlinear Model Predictive Controller with Scene Dynamics (DL-NMPC-SD) method for autonomous navigation. DL-NMPC-SD uses an \textit{a-priori} nominal vehicle model in combination with a scene dynamics model learned from temporal range sensing information. The scene dynamics model is responsible for estimating the desired vehicle trajectory, as well as to adjust the true system model used by the underlying model predictive controller. We propose to encode the scene dynamics model within the layers of a deep neural network, which acts as a nonlinear approximator for the high order state-space of the operating conditions. The model is learned based on temporal sequences of range sensing observations and system states, both integrated by an Augmented Memory component. We use Inverse Reinforcement Learning and the Bellman optimality principle to train our learning controller with a modified version of the Deep Q-Learning algorithm, enabling us to estimate the desired state trajectory as an optimal action-value function. We have evaluated DL-NMPC-SD against the baseline Dynamic Window Approach (DWA), as well as against two state-of-the-art End2End and reinforcement learning methods, respectively. The performance has been measured in three experiments: \textit{i}) in our GridSim virtual environment, \textit{ii}) on indoor and outdoor navigation tasks using our RovisLab AMTU (Autonomous Mobile Test Unit) platform and \textit{iii}) on a full scale autonomous test vehicle driving on public roads.

\end{abstract}


%
\IEEEpeerreviewmaketitle

\section{Introduction}
\label{sec:introduction}



\IEEEPARstart{I}{n} the last couple of years, Autonomous Vehicles (AVs) and self-driving cars began to migrate from laboratory development and testing conditions to driving on public roads. Their deployment in our environmental landscape offers a decrease in road accidents and traffic congestions, as well as an improvement of our mobility in overcrowded cities. Some of the main drivers behind the self-driving revolution are the advances in artificial intelligence and deep learning~\cite{Grigorescu_JFR_2019}. However, full autonomous driving is still an unsolved problem, with challenges ranging from better perception algorithms to improved motion controllers.

\begin{figure}
	\centering
	\begin{center}
		\includegraphics[scale=0.75]{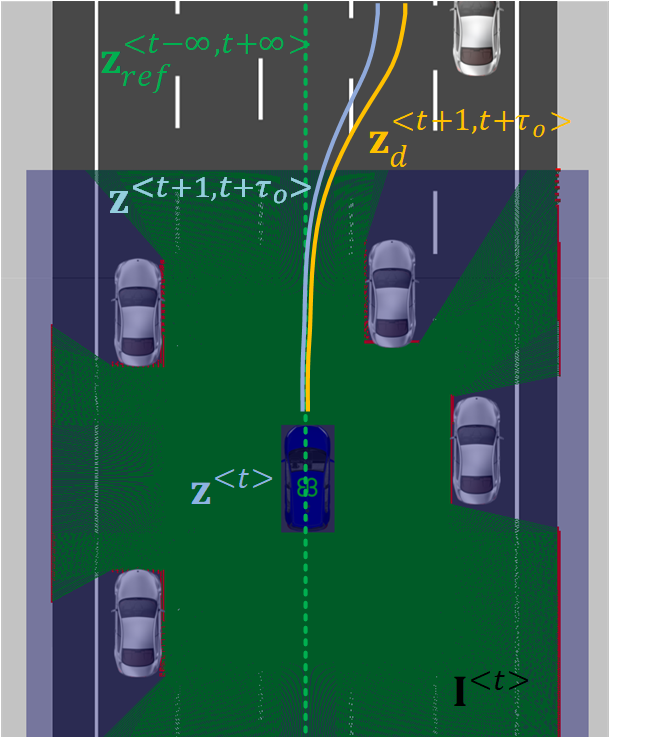}
		\vspace{-0.75em}
		\caption{\textbf{Autonomous driving problem space in the GridSim simulator~\cite{Trasnea_IRC2019}.} Given current and past vehicle states $\vec{z}^{<t-\tau_i, t>}$, a reference trajectory $\vec{z}_{ref}^{<t-\infty, t+\infty>}$ and an input sequence of observations $\vec{I}^{<t-\tau_i, t>}$, the goal is to estimate the vehicle's future state trajectory $\vec{z}^{<t+1, t+\tau_o>}$ and optimal control actions $\vec{u}^{<t+1, t+\tau_o>}$, over control horizon $\tau_o$.}
        \label{fig:problem_description}
	\end{center}
	\vspace{-1.5em}
\end{figure}

An autonomous vehicle, such as the one illustrated in Fig.~\ref{fig:problem_description}, is an intelligent agent which observes its environment, makes decisions and performs actions based on these decisions. The driving functions map sensory input to control output and are implemented either as modular perception-planning-action pipelines~\cite{Grigorescu_JFR_2019}, End2End~\cite{10614862_end2end} or Deep Reinforcement Learning (DRL)~\cite{9904958_DRL_Survey, 9537731_inverse_RL, das21a_inverse_RL} systems which directly map observations to driving commands. In a modular pipeline, the main problem is divided into smaller sub-problems, where each module is designed to solve a specific task and deliver the outcome as input to the adjoining component. Currently, the best self-driving performance is still offered by the modular pipelines, while End2End and DRL approaches are seeing significant breakthroughs in research.

Deep learning has become a leading technology in many domains, enabling AVs to perceive their driving environment and act accordingly. The current solutions for autonomous driving are typically based on machine learning concepts, which exploit large training databases acquired in different driving conditions~\cite{nuscenes_2020, Sun_2020_CVPR_waymo_DB, Kan_2024_icra_waymo_DB}. In a modular pipeline, deep learning is mainly used for perception. The detected and recognized objects are further passed to a path planner which calculates the reference trajectory for the AV's motion controller. The motion controller uses an a-priori vehicle model, together with the reference trajectory calculated by a path planner, to control the longitudinal and lateral velocities of the car. In such a system, if one component fails (e.g. path planning), the entire control system will produce erroneous outcomes.

Learning controllers in autonomous vehicles represent a middle ground between classical control systems and End2End learning-based approaches. They primarily involve classical control algorithms, like Model Predictive Control, where the controller's parameters are adjusted based on learning from training data rather than being fixed. Such an approach used in autonomous vehicles is the iterative single critic learning framework~\cite{9346029_Single_critic_LC_2021} which can effectively balance the frequency/changes in adjusting the vehicle’s control during the running process, or the Gaussian Process (GP) modelling~\cite{wang-jfr23, ostafew-ijrr16}, used for learning disturbance models.

In contrast to modular pipelines or learning controllers, End2End~\cite{10614862_end2end} and Deep Reinforcement Learning~\cite{9904958_DRL_Survey} systems are mostly model-free approaches, where driving commands are estimated directly from sensory information. Although the later systems perform better in the presence of uncertainties, they lack the predictable behavior which a model-based approach can offer. The stability is typically investigated in the sense of the learning algorithm’s convergence and not in the overall closed-loop stability principles.

Traditional NMPC controllers use an internal nominal vehicle model to predict future vehicle states. These predicted states are used to compute optimal control actions. However, such a standard approach does not take into account the external dynamics of the driving scene, typically observed using perception algorithms. In this work, we propose a method which follows a different paradigm, coined \textit{Deep Learning-based Nonlinear Model Predictive Controller with Scene Dynamics} (DL-NMPC-SD). It is based on a learning control synergy between a constrained NMPC and a joint internal \textit{a-priori} nominal vehicle model and a learned scene dynamics model computed from temporal range sensing data in an Inverse Reinforcement Learning (IRL) setup. The integration of the nominal model with the scene dynamics model aims to enable the computation of optimal control actions that also account for the dynamics of the driving environment. This approach allows us to predict how the vehicle's motion will be influenced by both the internal vehicle parameters and the anticipated changes in the scene (environment) along a receding horizon.

This approach mirrors human driving behavior, where speed and steering adjustments are made based on predictions of how the driving scene will unfold. It anticipates the vehicle's future state by not only considering its past position and velocity but also taking into account the dynamics of the driving environment, such as the movements of other traffic participants.

To implement DL-NMPC-SD, we utilize the concept of Augmented Memory, introduced in~\cite{SG_CyberCortex_2024} as a \textit{Temporal Addressable Memory}. The novelty of this memory type is its ability to store and access data based on the timestamp of its creation. The key advantage of using the Augmented Memory over other storage methods is its capability to access data collected from various sensors over a receding horizon, even when each sensor operates at different sampling frequencies. This enables the system to maintain separate sets of states, contol inputs and sensory data. In conjunction with a deep neural network, it is utilized to retrieve multimodal input data over a past temporal window (past receding horizon). This data is then propagated through the network layers to generate the output, which, in our case, represents the output of the scene dynamics model.

The joint dynamics model in our work is used to determine the optimal desired state trajectory of the vehicle, as well as to calculate a compensating control input added to the a-priori vehicle system model. We use temporal range sensing data to learn the scene dynamics model within a deep network trained in an inverse reinforcement learning setup, with a modified version of the Q-learning algorithm.

The above formulation exploits in a natural way the advantages of model-based control with the robustness of deep learning, enabling us to encapsulate the temporal dynamics of the driving scene within the layers of the deep network. The key contributions of the paper are:

\begin{itemize}
	\item the autonomous driving DL-NMPC-SD controller, based on an a-priori process model and a scene dynamics model;
	\item the nonlinear range sensing scene dynamics approximator for estimating the vehicle's desired trajectory and the compensating control input, encoded within the layers of a Deep Neural Network;
	\item a method for training the DL-NMPC-SD controller, based on Inverse Reinforcement Learning and a modified version of the Deep Q-Learning algorithm;
	\item training and evaluation of DL-NMPC-SD in our GridSim~\cite{Trasnea_IRC2019} virtual environment;
	\item indoor and outdoor navigation experiments on RovisLab AMTU (Autonomous Mobile Test Unit), as well as on public roads using a full scale autonomous test car.
\end{itemize}


The rest of the paper is organized as follows. Section~\ref{sec:related_work} covers the related work. The observations and the proposed process model are presented in Section~\ref{sec:method}, while the IRL training procedure is detailed in Section~\ref{sec:dynamics_model}. The experimental validation is given in Section~\ref{sec:experiments}. Finally, before the conclusions from Section~\ref{sec:conclusions}, we discuss the advantages and limitations of DL-NMPC-SD in Section~\ref{sec:discussion}. The integration of the process model in the NMPC controller is described in Appendix~A.

\section{Related Work}
\label{sec:related_work}


In the following, we provide an overview of model-free methods used for controlling autonomous vehicles, followed by a brief review of model predictive control and learning controllers applied to autonomous vehicles, and the state-of-the-art in autonomous navigation.

\subsection{Deep Learning based Control Methods}

Recent years have witnessed a growing trend in applying deep learning techniques to autonomous driving. These techniques are also encountered under the name of model-free methods, since they usually do not make use of an a-priori model of the underling controlled process~\cite{9904958_DRL_Survey}.

Among model-free approaches, the broad classes of \textit{End2End} and \textit{Deep Reinforcement Learning} systems are recognized today as some of the most promising approaches for enabling autonomous control in self-driving cars.

\textit{End2end learning}~\cite{10614862_end2end} directly maps raw input data to control signals. The training data, often in the form of images from a front-facing camera, is collected together with time-synchronized steering angles recorded from a human driver. A deep neural network is then trained by providing steering commands and input images of the road ahead. This type of approach has been backed by new automotive training databases, such as NuScenes~\cite{nuscenes_2020}, or te Waymo datasets~\cite{Sun_2020_CVPR_waymo_DB, Kan_2024_icra_waymo_DB}.

Our task is similar to the aggressive driving task considered in~\cite{Pan_IJRR_19}. Compared with DL-NMPC-SD, their deep learning policy is trained for agile driving on a predefined obstacles-free track, where steering and throttle commands are predicted from images in an End2End fashion. This approach limits the applicability of their system to autonomous driving, since a self-driving car has to navigate roads with dynamic obstacles and undefined lane boundaries.

End2end systems are faced with the challenge of learning a very complex mapping in a single step. In~\cite{XuGYD16}, a combined convolutional and recurrent network is proposed for the end2end learning of a driver model. The convolutional network processes single instances of images, while the recurrent net integrates the moving path of the vehicle. Although end2end approaches behave well in certain low-speed situations, a high capacity model together with a large amount of training data is required to learn corner cases with multi-path situations, such as a T-point or road intersections.

\textit{Deep Reinforcement Learning} (DRL)~\cite{mnih2015humanlevel, 9904958_DRL_Survey} is a type of machine learning algorithm where agents are taught actions by interacting with their environment. In such a system, a policy is a mapping from a state to a distribution over actions. The algorithm does not leverage on training data, but maximizes a cumulative reward which is positive if the vehicle is able to maintain its direction without collisions and negative otherwise. The reward is used as a pseudo label for training a deep neural network, which is then used to estimate an action-value function approximating the next best driving action, given the current state. This is in contrast with End2End learning, where labeled training data is provided. Since the agent has to explore its environment, the training is usually performed through learning from collisions.

DRL has been successfully used to learn complex mobile manipulation skills on physical robots. Guided policy search~\cite{Levine_2015_ICRA, 9537641_RL_nav} uses a neural network to learn a global policy from simple local policies computed on the robot. In~\cite{LevinePKIQ2018, 9199280_DL_in_robotics_2021}, many physical robots have been used in parallel to collect sufficient experience for training a DRL system for mobile manipulation. A barrier Lyapunov function based on reinforcement learning is proposed in~\cite{9649902_Safety_RL} in the context of safety in autonomous driving.

Although several methods have been proposed for improving model-free DRL, such as the soft actor-critic in~\cite{pmlr-v80-haarnoja18b}, the main challenge with DRL on physical systems is the very high sample complexity and brittle convergence properties, as well as the need for the agent to explore its working environment. This is challenging in autonomous driving, since traffic conditions cannot be fully replicated in a controlled laboratory setup, as in the case of mobile manipulation. A solution here is provided by \textit{Inverse Reinforcement Learning} (IRL), which is an imitation learning method for solving Markov Decision Processes (MDPs). A representative IRL approach is the Maximum Entropy IRL~\cite{ZiebartMBD08}, used to optimize the MDP's objective by fitting a cost function from a family of cost functions. In robotic navigation, IRL offers a solution for learning navigation policies~\cite{9904958_DRL_Survey, 9537731_inverse_RL, das21a_inverse_RL}. In~\cite{WulfmeierWP16, BarnesMP16}, Maximum Entropy IRL has been extended with a fully convolutional deep neural network for learning a navigation cost map in urban environments. However, such methods usually do not take into account the vehicle's state and the feedback loop required for low-level control, providing instead only a reference set-point to a low-level controller.

\subsection{Model Predictive Control}

\textit{Model Predictive Control} (MPC)~\cite{GARCIA_Automatica_1989} is a control strategy that computes control actions by solving an optimization problem. It received lots of attention in the last two decades due to its ability to handle complex nonlinear systems with state and input constraints. The central idea behind MPC is to calculate control actions at each sampling time by minimizing a cost function over a short time horizon, while considering observations, input-output constraints and the system's dynamics given by a process model~\cite{rawlings2009model}.

MPC has proved to be a reliable control technique for self-driving cars~\cite{BBoots_Aggressive_Driving_2024}, autonomous mobile robots~\cite{wang-jfr23} and unmanned aerial vehicles~\cite{krinner_RSS_2024}. This is mainly due to the fact that MPC is able to incorporate complex mathematical models that take into account the dynamics of the modeled vehicle~\cite{BBoots_Aggressive_Driving_2024}. Unconstrained MPC based on linearized kinematic models has been used in vehicle controllers~\cite{MPC_path_tracking_survey_2023}, as well as for path and trajectory tracking~\cite{MPC_path_tracking_survey_2023}.

\subsection{Learning Controllers}

Traditional controllers make use of an \textit{a-priori} model composed of fixed parameters. When robots or other autonomous systems are deployed in complex environments, such as driving, traditional controllers cannot foresee every possible situation that the system has to cope with. Unlike controllers with fixed parameters, learning controllers make use of training information to learn their models over time. With every gathered batch of training data, the approximation of the true system model becomes more accurate, thus enabling both model flexibility and consistent uncertainty estimates~\cite{Rasmussen2006}.

In previous works, learning controllers have been introduced based on simple function approximators, such as Gaussian Process (GP) modeling~\cite{wang-jfr23, ostafew-ijrr16}, or Support Vector Regression~\cite{Sigaud_2011}. Learning based unconstrained and constrained NMPC controllers have been proposed in~\cite{wang-jfr23} and~\cite{ostafew-ijrr16}, respectively. Given a trajectory path, both approaches use a simple \textit{a-priori} model and a GP learned disturbance model for the path tracking control of a mobile robot. The experimental environment comprises of fixed obstacles, without any moving objects. Unlike these examples, in our work we model the dynamics of the driving scene within the layers of a deep neural network.

Neural networks have been previously used to model plant dynamics~\cite{ARNOLD2021104195}. In the last years, deep generative models based on neural networks have been applied to model-based control of redundant robotic manipulators based on visual data~\cite{JPeters_survey_2024}. In~\cite{9684679_Gait}, a medium size deep network is combined with MPC and reinforcement learning to produce stable gaits and locomotion tasks.

\subsection{Autonomous Navigation}

Recent research on DRL and its application to autonomous navigation has explored various strategies to enhance safety, performance, and efficiency in complex environments. A Barrier Lyapunov Function-based safe RL (BLF-SRL) approach~\cite{9649902_Safety_RL} has been proposed to ensure safe exploration by constraining state variables within a designed safety region during learning, effectively reducing control performance variance under uncertainty. Complementary to this, Deductive Reinforcement Learning (DeRL)~\cite{9537641_RL_nav} incorporates a deduction reasoner for vision-based autonomous urban driving, allowing the agent to predict future transitions and assess the consequences of current policies, leading to safer and more reliable decision-making. Deep Q-Network (DQN) and Deep Deterministic Policy Gradient (DDPG) have also been implemented to control autonomous vehicles within the Carla simulator~\cite{Perez_Carla_2022}, showing that both algorithms can successfully navigate routes. Additionally, hierarchical deep learning frameworks combining motion planning with DRL-based collision-free control have been developed, enhancing autonomous exploration by improving motion planning performance and reducing training time through techniques like noisy prioritized experience replay~\cite{9913936_Mobile_Robot_Nav_2024}. Despite these advances, challenges remain in the form of sample inefficiency and exploration in complex environments, as highlighted in comprehensive surveys on exploration methods in single-agent and multi-agent RL~\cite{9904958_DRL_Survey, 10021988_survey_DRL_2024}.

Although promising, these approaches have been tested mainly in simulation environments or in simplified settings designed for small-scale robots~\cite{9913936_Mobile_Robot_Nav_2024}. We believe that the primary challenge is the "curse of dimensionality" of the scene's state-space, as RGB images often include a substantial amount of information. In our approach, we tackle this issue by reducing the scene's state-space dimension through the use of real-world range sensing observations rather than RGB data, and we perform training and testing in real-world autonomous navigation and driving scenarios.

\begin{table}
	\centering
	\begin{tabular}{ll}
		\hline
			Symbol & Description \\ 
		\hline
			\multicolumn{2}{l}{\textbf{Generic definitions}} \\
				$t$ & Discrete time \\
				$\Delta t$ & Sampling time \\
				$\tau_i$ & Past (input) temporal horizon \\
				$\tau_o$ & Future (output) temporal horizon \\
				$[t-\tau_i, t]$ & Past receding horizon (window) \\
				$[t+1, t+\tau_o]$ & Prediction receding horizon (window) \\
				$\vec{p} = (x, y)$ & Vehicle position (birds-eye view) \\
				$\rho$ & Vehicle heading (yaw) \\
			\\
			\multicolumn{2}{l}{\textbf{Modelling and control variables}} \\
				$\vec{f}(\cdot)$ & Nonlinear vehicle model \\
				$\vec{h(\cdot)}$ & Scene dynamics model \\
				$\vec{z}^{<t>}$ & Vehicle state at time $t$ \\
				$\vec{s}^{<t>} \in S$ & Joint vehicle and environment state \\
				& (past vehicle states and past observations) \\
				$v_{cmd}$ & Longitudinal velocity control input \\
				$\delta_{cmd}$ & Steering angle control input \\
				$\vec{u}^{<t>}$ & Control actions at time $t$ \\
				$J(\vec{z}, \vec{u})$ & Cost function over receeding horizon \\
				$\vec{Q}, \vec{R}$ & Symmetric cost matrices in $J(\vec{z}, \vec{u})$ \\
			\\		
			\multicolumn{2}{l}{\textbf{Temporal sequence data}} \\
				$\mathcal{D}$ & Dataset of training sequences \\
				$\vec{z}^{<t-\infty, t+\infty>}_{ref}$ & Global reference trajectory \\
				$\vec{z}^{<t-\tau_i, t>}$ & Past vehicle states \\
				$\vec{z}^{<t+1, t+\tau_o>}_d$ & Future (desired) vehicle state trajectory \\
				$\vec{I}^{<t-\tau_i, t>}_t$ & Temporal sequence of observations \\
				$\vec{u}^{<t+1, t+\tau_o>}$ & Control actions over prediction horizon \\
			\\
			\multicolumn{2}{l}{\textbf{MDP and Q-learning formulation}} \\
				$M$ & Markov Decision Process \\
				$S$ & Finite set of states \\
				$Z_d$ & Finite set of set-point sequences \\
				$\mathcal{T}$ & Stochastic transition function \\
				$R$ & Scalar reward function \\
				$\vec{w}$ & Reward function weights \\
				$\gamma$ & Discount factor \\
				$Q(\cdot, \cdot)$ & Action-value function \\
				$\pi$ & Behavioral policy (action) \\
				$\Theta$ & Deep neural network weights \\
				$\nabla_{\Theta}$ & Gradients within the deep neural network \\
				$l(\cdot, \cdot)$ & Logistic regression cost function \\
				$\vec{e}^{<t>}$ & Cross-track error constraint \\
			\\
			\multicolumn{2}{l}{\textbf{Evaluation metrics}} \\
				$e_L$ & Cumulative speed-weighted lateral error \\
				$e_H$ & Cumulative speed-weighted heading error \\
		\hline
	\end{tabular}
	\caption{Notations and variables used throughout the paper.}
	\label{tab:notations}
	\vspace{-2.5em}
\end{table}

\vspace{1em}

In the light of the current approaches and their limitations, we propose the DL-NMPC-SD method, where we leverage on a low-level constrained NMPC controller tightly coupled to a learned scene dynamics model. Based on temporal range sensing data, the model computes the desired trajectory in a closed-loop manner, while predicting the vehicle's future states required by the NMPC controller.

\section{Observations and Process Modelling}
\label{sec:method}

MPC~\cite{MPC_path_tracking_survey_2023, rawlings2009model} and Reinforcement Learning~\cite{9904958_DRL_Survey, Sutton_RL} are both methods for optimal control of dynamic systems which have evolved in parallel in the control systems and computational intelligence communities, respectively. An overview of the relations between the two paradigms is given in~\cite{GORGES20174920}. Throughout the paper, we use the following notation. The value of a variable is defined either for a single discrete time step $t$, written as superscript $<t>$, or as a discrete sequence defined in the $<t, t+k>$ time interval, where $k$ denotes the length of the sequence. For example, the value of a state variable $\vec{z}$ is defined either at discrete time $t$ as $\vec{z}^{<t>}$, either within a sequence interval $\vec{z}^{<t, t+k>}$. Vector and matrices are indicated by bold symbols.

The variables used througout the paper are listed in Table~\ref{tab:notations}.

\subsection{Problem Definition}


A basic illustration of the autonomous driving problem is shown in Fig.~\ref{fig:problem_description}. Given a global reference trajectory route $\vec{z}^{<t-\infty, t+\infty>}_{ref}$, a temporal sequence of observations $\vec{I}^{<t-\tau_i, t>}$ and past vehicle states $\vec{z}^{<t-\tau_i, t>}$, we want to estimate the optimal control actions $\vec{u}^{<t+1, t+\tau_o>}$ over receding horizon $\tau_o$, such that the vehicle follows a desired future state trajectory $\vec{z}^{<t+1, t+\tau_o>}_d$. Since the desired trajectory is the input to the NMPC, we will also refer to $\vec{z}^{<\cdot>}_d$ as set-point. Considering a discrete time $t$, we define $\tau_i$ and $\tau_o$ as past and future temporal horizons, respectively.

The vehicle is modeled based on the kinematic bicycle model of a robot~\cite{BBoots_Aggressive_Driving_2024}, with position state $\vec{z}^{<t>} = (x^{<t>}, y^{<t>}, \rho^{<t>})$ and no-slip assumptions. $x$, $y$ and $\rho$ represent the position and heading of the vehicle in the 2D driving plane, respectively. We apply the longitudinal velocity and the steering angle as control actions: $\vec{u}^{<t>} = (v^{<t>}_{cmd}, \delta^{<t>}_{cmd})$.

The reference trajectory $\vec{z}_{ref}$ represents the global route which the vehicle should follow, from its start position to destination, given as a set of waypoints (e.g. GPS coordinates). Since $\vec{z}_{ref}$ is a global trajectory, we consider it to vary in the $(t-\infty, t+\infty)$ interval. For practical reasons, the global reference trajectory is stored at sampling time $t$ over a finite horizon $[t - \tau_i, t + \tau_o]$. As shown in Section~\ref{sec:rcnmpc}, $\vec{z}_{ref}$ is different from the desired trajectory $\vec{z}^{<t+1, t+\tau_o>}_d$, which is dependent on the scene's dynamics and is calculated over a finite horizon $[t+1, t+\tau_o]$. $\vec{z}^{<\cdot>}_d$ is executed by the NMPC controller and must take into account the drivable area and the obstacles present in the scene.

In our work, we seek to develop a learned scene dynamics model that can be integrated with the vehicle's nominal process model to achieve two objectives: \textit{i}) to compute a desired vehicle trajectory $\vec{z}^{<\cdot>}_d$ over a receding horizon, and \textit{ii}) to determine the optimal control actions $\vec{u}^{<t>}$ that will safely track $\vec{z}^{<\cdot>}_d$.

\subsection{DL-NMPC-SD Overview}
\label{sec:overview}

\begin{figure}
	\centering
	\begin{center}
		\includegraphics[scale=0.9]{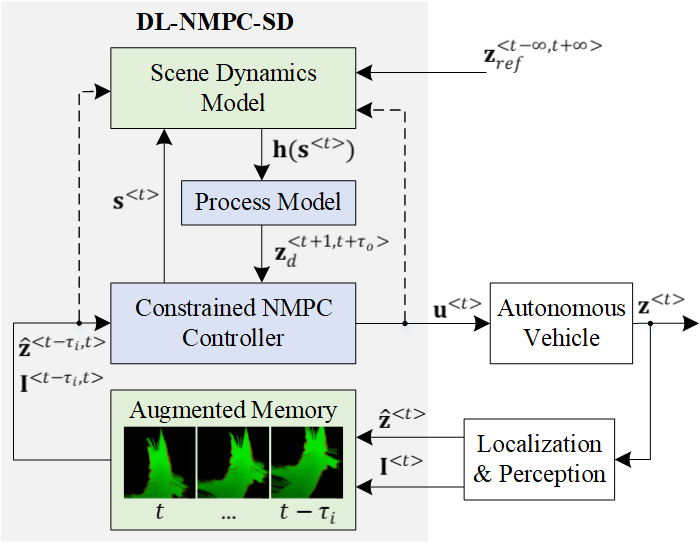}
		\vspace{-1em}
		\caption{\textbf{DL-NMPC-SD: Deep Learning-based Nonlinear Model Predictive Controller using Scene Dynamics for autonomous navigation}. The desired trajectory $\vec{z}^{<t+1, t+\tau_o>}$ for the constrained NMPC is estimated by the learned scene dynamics model $\vec{h}(\cdot)$ added to the nominal process model, based on historical sequences $\vec{s}^{<t>} = (\vec{z}^{<t-\tau_i, t>}, \vec{I}^{<t-\tau_i, t>})$ of system states $\vec{z}^{<t-\tau_i, t>}$ and observations $\vec{I}^{<t-\tau_i, t>}$, each integrated by the Augmented Memory component. The input required by the controller is the global start-to-destination route $\vec{z}^{<t-\infty, t+\infty>}_{ref}$ for the vehicle, which, from a control perspective, varies in the $(t-\infty, t+\infty)$ interval. $\tau_i$ and $\tau_o$ represent past (input) and future (output) temporal horizons, respectively. The dotted lines illustrate the flow of data used during training.}
        \label{fig:block_diagram}
	\end{center}
	\vspace{-1.5em}
\end{figure}

The block diagram of DL-NMPC-SD is shown in Fig.~\ref{fig:block_diagram}. At its core, the approach uses a process model (Section~\ref{sec:rcnmpc}), defined as a combination of a known nominal vehicle model $\vec{f}(\vec{z}^{<t>}, \vec{u}^{<t>})$ and a learned scene dynamics model $\vec{h}(\vec{s}^{<t>})$. Using a past (receding) temporal horizon, the process model is used to predict the future state of the vehicle based on its past states (nominal model) $\vec{\hat{z}}^{<t-\tau_i, t>}$ and based on the dynamics of the observed scene (scene dynamics model). The observations are composed of sequences of occupancy grids $\vec{I}^{<t-\tau_i, t>}$ (Section~\ref{sec:perception}), while the scene dynamics model is encoded within the layers of a deep neural network (Section~\ref{sec:dnn}).

Since the inputs and outputs are temporal sequences, we leverage on the \textit{Temporal Addressable Memory} concept introduced in~\cite{SG_CyberCortex_2024}, which is implemented in this work as the Augmeted Memory from Fig.~\ref{fig:block_diagram}. The Augmented Memory acts as a buffer for storing data that can be accessed over a receding past temporal interval $[t-\tau_i, t]$. This data is subsequently propagated through the layers of the deep network.

In order to use the scene dynamics model, we first have to train it on a dataset expert actions (Section~\ref{sec:dynamics_model}). To enhance generalization, we employ a two-step Inverse Reinforcement Learning approach: 1) first, we learn the linear coefficients of the reward function, 2) secondly, we used the learned reward to adapt the weights of the deep neural network using a modified version of the Deep Q-Learning algorithm.

The final component of DL-NMPC-SD is the constrained NMPC controller, which computes the control input $\vec{u}^{<t>}$ through an iterative optimization procedure (Appendix~A). The accuracy of this optimization process depends on the underlying dynamics model. In this context, the scene dynamics model serves two key functions: first, it provides a desired state trajectory $\vec{z}^{<t+1,t+\tau_o>}_d$ for the controller to follow; second, it is used to accurately predict the vehicle's state within the optimization loop of the NMPC algorithm, enabling the computation of $\vec{u}^{<t>}$.

\subsection{Driving Scene Perception using Range Sensing}
\label{sec:perception}

\begin{figure*}
	\centering
	\begin{center}
		\includegraphics[scale=0.9]{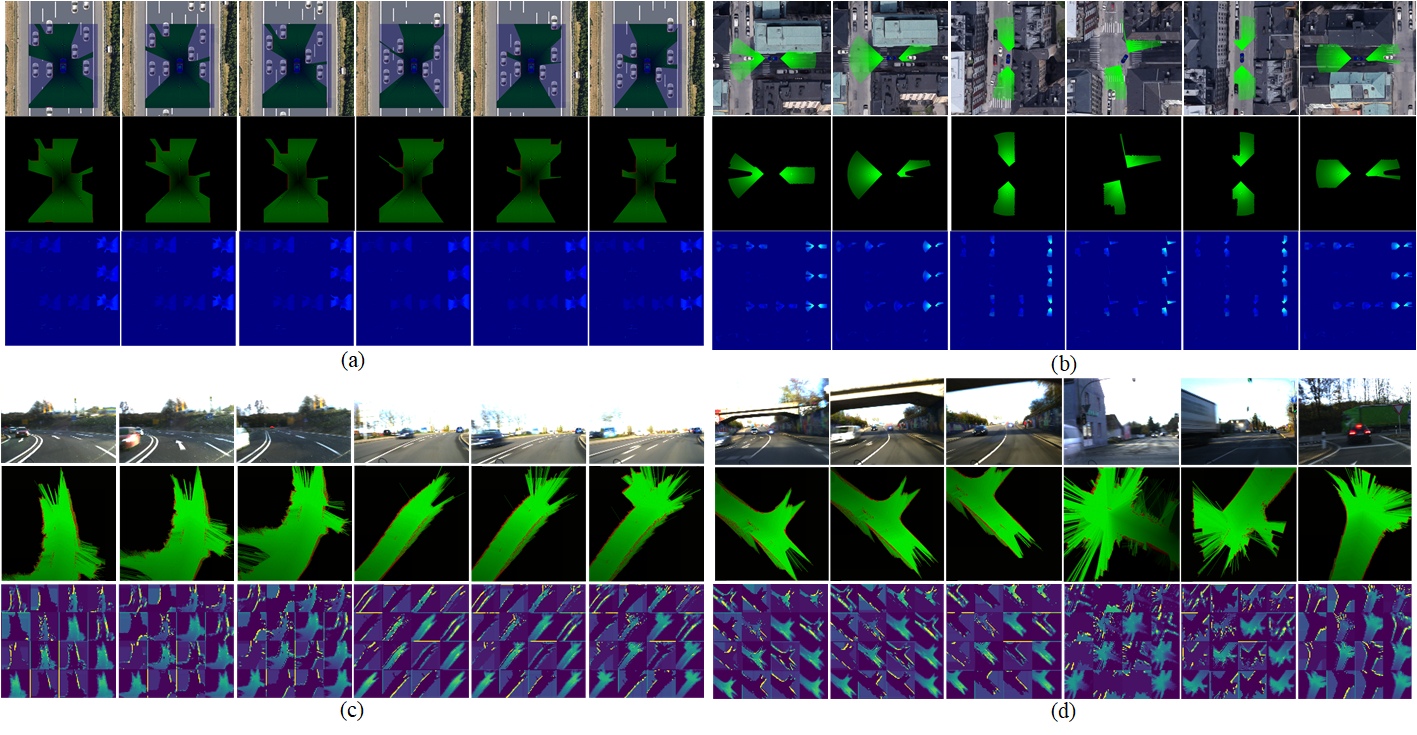}
			\vspace{-1em}
		\caption{\textbf{Examples of synthetic GridSim (a,b) and real-world (c,d) occupancy grids.} The images in each group show visual snapshots of the driving environment together with their respective OGs and activations of the first convolutional layer of the network from Fig.~\ref{fig:neural_network_diagram}.}
        \label{fig:occupancy_grids}
	\end{center}
	\vspace{-1.5em}
\end{figure*}

We observe the driving environment using LiDAR and radar data fused as 2D \textit{Occupancy Grids} (OGs). In our previous work, we have leveraged on OGs and deep networks to estimate the optimal future states of a vehicle~\cite{Grigorescu_RAL_2019}, as well as the driving context in which a self-driving car navigates~\cite{Marina_IRC2019}. A single OG corresponds to an observation instance $\vec{I}^{<t>}$, while a sequence of OGs is denoted as $\vec{I}^{<t-\tau_i, t>}$. These observations are axis-aligned discrete grid sequences, acquired over time interval $[t-\tau_i, t]$ and centered on the vehicle's states $\vec{z}^{<t-\tau, t>}$.

Occupancy grids provide a birds-eye perspective on the traffic scene, as shown in the examples from Fig.~\ref{fig:occupancy_grids}. The basic idea behind OGs is the division of the environment into 2D cells, each cell representing the probability, or belief, of occupation through color-codes. Green represents free space, red pixels show occupied cells (or obstacles) and black signifies an unknown occupancy. The intensity of the color represents the degree of occupancy. The higher the intensity of the green color is, the higher the probability of a cell to be free is.

We assume that the driving area should coincide with free space, while non-drivable areas may be represented by other traffic participants, road boundaries, buildings, or other obstacles. OGs are often used for environment perception and navigation. In our work, we have constructed OGs using the \textit{Dempster-Shafer} (DS) theory, also known as the \textit{Theory of Evidence} or the \textit{Theory of Belief Functions}~\cite{DS_Survey_2022}. Synthetic data has been generated in GridSim~\cite{Trasnea_IRC2019} based on an OG sensor model. The implementation details of OG based perception is given in Appendix~D.

The localization of the vehicle, that is, the computation of state estimate $\vec{z}^{<t>}$, is obtained by applying Kalman filtering on the wheel's odometry and the double integration of the acceleration acquired from an Inertial Measurement Unit (IMU).

\subsection{Process Model}
\label{sec:rcnmpc}

Consider the following nonlinear state-space system, describing the vehicle's transition from one state to another:

\begin{equation}
	\vec{z}^{<t+1>} = \vec{f}_{true} (\vec{z}^{<t>}, \vec{u}^{<t>}),
\end{equation}

\noindent with observable state $\vec{z}^{<t>} \sim \mathcal{N} (\vec{\bar{z}}^{<t>}, \sigma^2_f)$, $\vec{z}^{<t>} \in \mathbb{R}^n$, mean state $\vec{\bar{z}}^{<t>}$ and control input $\vec{u}^{<t>} \in \mathbb{R}^m$, at discrete time $t$. We assume that each state measurement is corrupted by zero-mean additive Gaussian noise with variance $\sigma^2_f$. The true system $\vec{f}_{true}$ is not known exactly and is approximated by a function depending on the \textit{a-priori} nominal process model $\vec{f(\cdot)}$ and the experience-based scene dynamics model $\vec{h(\cdot)}$:

\begin{equation}
	\vec{z}^{<t+1>} = \underbrace{\vec{f}(\vec{z}^{<t>}, \vec{u}^{<t>})}_{\text{nominal model}} + \underbrace{\vec{h}(\vec{s}^{<t>})}_{\text{scene dynamics model}},
	\label{eq:true_system_model}
\end{equation}

\noindent with environmental dependencies $\vec{s}^{<t>} \in \mathbb{R}^p$.

$\vec{s}^{<t>} = (\vec{z}^{<t-\tau_i, t>}, \vec{I}^{<t-\tau_i, t>})$ is defined at time $t$ as the joint vehicle states $\vec{z}^{<\cdot>}$ and the observed states of the environment (scene) $\vec{I}^{<\cdot>}$, both along past receding horizon $[t-\tau_i, t]$. The set of past temporal vehicle and scene states are integrated by the \textit{Augmented Memory} component, which stores states and observations from the past receding horizon $[t-\tau_i, t]$. It's novelty lies in its dual function as both an online and offline data storage and retrieval system, similar to the experience replay buffer used in Deep Q-learning. During online operation, the stored data is used to calculate the compensating action of the scene dynamics model $\vec{h}(\cdot)$. During trainnig (dotted lines in Fig.~\ref{fig:block_diagram}), it provides the Q-learning system with access to the trainnig data used for adapting the weights of the deep neural network.

The rationale for integrating the nominal \textit{a-priori} model $\vec{f}(\cdot)$ with the learned scene dynamics model $\vec{h}(\cdot)$ is to improve the prediction accuracy of future vehicle states. To compute optimal control actions over a future receding horizon $[t+1, t+\tau_o]$, we need to predict how the vehicle's motion will be influenced not only by the parameters of the nominal model, but also by the anticipated changes in the scene, within the considered receding interval.

The model $\vec{f}(\cdot)$ is the nonlinear process model, composed of the known component, representing our knowledge of $\vec{f}_{true}(\cdot)$, while $\vec{h}(\cdot)$ is the learned scene dynamics component, defined as:

\begin{equation}
    \vec{h}(\vec{s}^{<t>}) = 
        \begin{bmatrix} 
            h_{v}(\vec{s}^{<t>}) \\
            h_{\delta}(\vec{s}^{<t>})
        \end{bmatrix}
\end{equation}

\noindent where $h_v$ and $h_{\delta}$ are compensating factors calculated through the scene dynamics model, and added to the vehicle process model. With the sampling time defined as $\Delta t$, the process model employed by DL-NMPC-SD is:

\begin{equation}
	\vec{z}^{<t+1>} = \vec{z}^{<t>} + 
		\begin{bmatrix}
			\cos \rho^{<t>} + h_{v} \\
			\sin \rho^{<t>} + h_{v} \\
			 \frac{1}{L} \cdot \tan \delta^{<t>}_{cmd} \cdot \cos h_{\delta}
		\end{bmatrix}
		\Delta t
	\label{eq:nominal_process_model}
\end{equation}

\noindent where $L$ is the length between the front and the rear wheel.

As with any system identification method, the approximation of  $\vec{f}_{true} (\cdot)$ is subject to identification errors. Typically, identification errors are represented as disturbance terms that account for the discrepancy between the true function and the approximated model: $\vec{\hat{f}} = \vec{f}_{true} + d$, where $\vec{\hat{f}}$ is the approximated model and $d$ denotes the error term. The identification error quantifies the difference between the true model and the estimated models. Since both $\vec{f}(\cdot)$ and $\vec{h}(\cdot)$ differ from $\vec{f}_{true}$, they can be weighted according to their respective estimation errors. For clarity and to avoiding the weighting bias, we do not assign weights to the two models. Instead, we use the analytical nominal model without the disturbance term, while the identification error is not considered as a separate quantity in the case of the learned scene dynamics model $\vec{h}(\cdot)$. The identification errors will thus be encoded within the deep neural network estimator.

\subsection{Deep Neural Network Architecture for Scene Dynamics Modelling}
\label{sec:dnn}

Given a sequence of temporal observations $\vec{I}^{<t-\tau_i, t>}: \mathbb{R}^w \times \tau_i \rightarrow \mathbb{R}^w \times \tau_o$, the system's state $\vec{z}^{<t>} \in \mathbb{R}^n$ and the reference set-points $\vec{z}^{<t+\tau_o>}_{ref} \in \mathbb{R}^n$ in observation space at time $t$, the task is to learn a sequence of desired set-points $\vec{z}^{<t+1, t+\tau_o>}_d$ (driving strategy) for navigating from state $\vec{z}^{<t>}$ to the destination state $\vec{z}^{<t+\tau_o>}_{ref}$, where $\tau_o$ is the length of the receding horizon.

The output of the deep neural network encoding the scene dynamics model, is a sequence of desired vehicle states (set-points) $\vec{z}^{<t+1, t+\tau_o>}_d$:

\begin{equation}
	\vec{z}^{<t+1, t+\tau_o>}_d = [\vec{z}^{<t+1>}_d, \vec{z}^{<t+2>}_d, ..., \vec{z}^{<t+\tau_o>}_d],
	\label{eq:behavioral_policy}
\end{equation}

\noindent where $\vec{z}^{<t+i>}_d$ is a predicted vehicle state at time $t+i$, calculated according to Eq.~\ref{eq:nominal_process_model} . $\tau_i$ and $\tau_o$ are not necessarily equal input and output temporal horizons, namely $\tau_i \neq \tau_o$.

\begin{figure}
	\centering
	\begin{center}
		\includegraphics[scale=0.9]{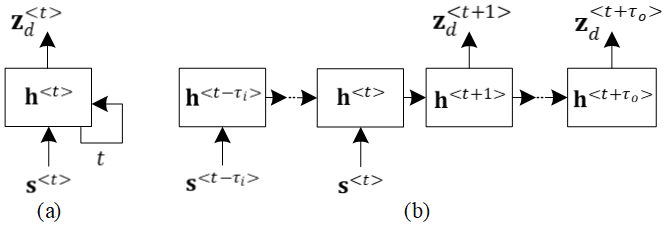}
		\vspace{-1em}
		\caption{\textbf{A folded (a) and unfolded (b) over time, many-to-many Recurrent Neural Network}. Both the input $\vec{s}^{<t-\tau_i, t>}$ and output $\vec{z}^{<t+1, t+\tau_o>}_d$ sequences share the same weights over time $t$.}
        \label{fig:rnn_block_diagram}
	\end{center}
	\vspace{-1.5em}
\end{figure}

In the last couple of years, deep learning has been established as the main technology behind many innovations, showing significant improvements in perception and navigation~\cite{9764831_DL_NAV_survey_2023}, robotics~\cite{9199280_DL_in_robotics_2021} and Large Language Models (LLMs)~\cite{openai2023gpt}. Among the deep learning techniques, Recurrent Neural Networks (RNN) are especially good in processing temporal sequence data, such as text, or video streams. Different from conventional neural networks, a RNN contains a time dependent feedback loop in its memory cell. Given a time dependent input sequence $[\vec{s}^{<t-\tau_i>}, ..., \vec{s}^{<t>}]$ and an output sequence $[\vec{z}^{<t+1>}_d, ..., \vec{z}^{<t+\tau_o>}_d]$, a RNN can be "unfolded" $\tau_i + \tau_o$ times to generate a loop-less network architecture matching the input length, as illustrated in Fig.~\ref{fig:rnn_block_diagram} . $t$ represents a temporal index, while $\tau_i$ and $\tau_o$ are the lengths of the input and output sequences, respectively. An unfolded network has $\tau_i + \tau_o + 1$ identical layers, where each layer shares the same learned weights. Once unfolded, a RNN can be trained using the backpropagation through time algorithm. When compared to a conventional neural network, the only difference is that the learned weights in each unfolded copy of the network are averaged, thus enabling the network to share the same weights over time.

\begin{figure*}
	\centering
	\begin{center}
		\includegraphics[scale=0.95]{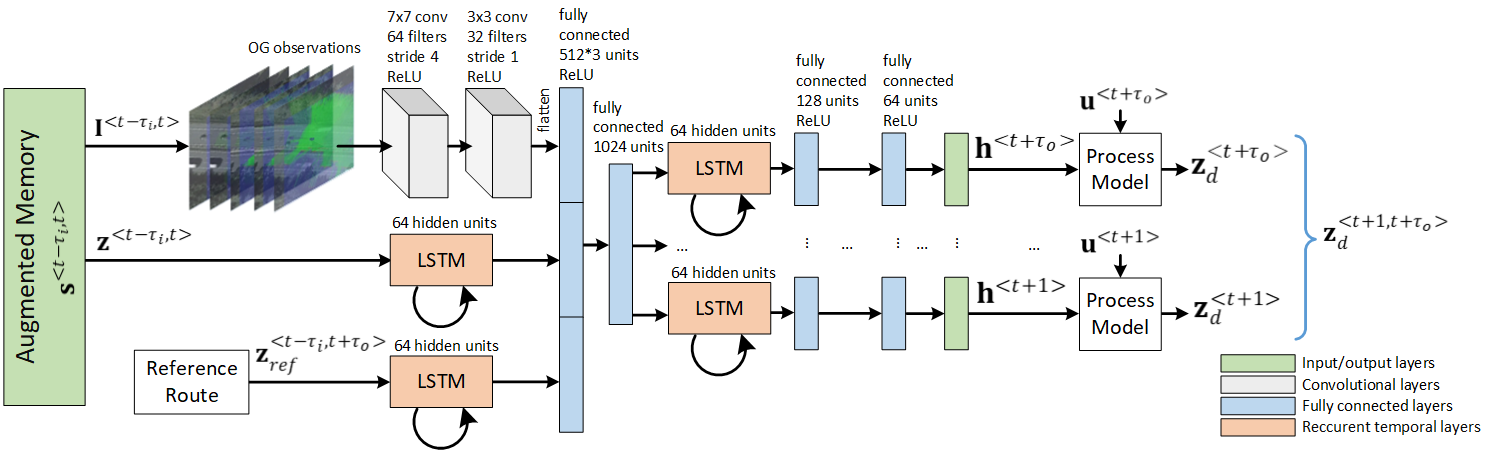}
		\vspace{-0.5em}
		\caption{\textbf{Deep Neural Network architecture for encoding the scene dynamics model $\vec{h}(\cdot)$. Combined with the nominal process model, the output of the network is used to calculate desired policies along receding horizon $\vec{z}^{<t+1, t+\tau_o>}_d$.} The training data consists of synthetic and real-world observation sequences $\vec{I}^{<t-\tau_i, t>}$, vehicle states $\vec{z}^{<t-\tau_i, t>}$, and reference trajectories $\vec{z}^{<t-\tau_i, t+\tau_o>}_{ref}$, together with their respective desired states $\vec{z}^{<t+1, t+\tau_o>}_d$. Before being stacked upon each other, the observations stream is passed through a convolutional neural network, having two convolutional blocks of $9.472$ and $896$ units, respectively. The input vehicle and reference states are firstly process by LSTM modules, each composed of $64$ hidden units. The merged data is further fed to separate LSTMs via two fully connected layers of $1536$ and $1024$ units, respectively. Each LSTM branch is responsible for predicting a compensating factor $\vec{h}^{<t+i>}$, where $i$ is an index in the receding horizon.}
        \label{fig:neural_network_diagram}
	\end{center}
	\vspace{-1.5em}
\end{figure*}

In our work, we use a set of Long Short-Term Memory (LSTM)~\cite{hochreiter1997long} networks, which act as non-linear function approximators for estimating temporal dependencies in observation sequences (occupancy grids).

The deep neural network architecure for estimating the scene dynamics model based on temporal sequence data, is illustrated in Fig.~\ref{fig:neural_network_diagram}., and is parametrized by its network weights $\Theta$.

In the deep network from Fig.~\ref{fig:neural_network_diagram}, the OG observations are firstly passed through two blocks of convolutional, ReLU activations and max pooling layers. The first block is composed of $64$ convolutional filters of size $7 \times 7$ and stride 4, while the second block has $32$ filters of size $3 \times 3$ and stride 1. Overall, the convolutional blocks consist of $9.472$ and $896$ units (neurons), respectively. This builds an abstract representation of the observations, which is stacked on top of embeddings of previous vehicle states $\vec{z}^{<t-\tau_i, t>}$ and reference trajectory $\vec{z}^{<t-\tau_i, t+\tau_o>}_{ref}$, obtained using two LSTM modules, each having $64$ hidden units. The stacked representation is analyzed by a series of LSTM branches, where each branch is responsible for estimating a desired set-points trajectory along receding horizon $[t+1, t+\tau_o]$. Each LSTM branch has $64$ hidden neurons and is designed to determine the compensating scene dynamics term (Eq.~\ref{eq:true_system_model}) for a specific sampling time in the receding horizon.

The process model component from Fig.~\ref{fig:neural_network_diagram} is used to combine the nominal process model with the scene dynamic model, according to Eq.~\ref{eq:true_system_model}. $\vec{u}^{<\cdot>}$ is an input, since the nominal process model function requires it in order to analytically predict the next states, as shown in Eq.~\ref{eq:nominal_process_model}.

In a supervised learning setup, also refered to as Behavioral Cloning, given a set of training sequences $\mathcal{D} = [(\vec{s}^{<t-\tau_i, t>}_1, \vec{z}^{<t+1, t+\tau_o>}_{d_1}), ..., (\vec{s}^{<t-\tau_i, t>}_q, \vec{z}^{<t+1, t+\tau_o>}_{d_q})]$, that is, $q$ independent pairs of observed sequences with assignments $\vec{z}^{<t, t+\tau_o>}_d$, one can train the response of a neural network $Q(\cdot; \Theta)$ using Maximum Likelihood Estimation (MLE):

\begin{equation}
	\begin{split}
		\hat{\Theta} & = \arg \max_{\Theta} \mathcal{L} (\Theta; \mathcal{D}) \\
		& = \arg \min_{\Theta} \sum_{i=1}^m l_i (Q (\vec{s}^{<t-\tau_i, t>}_i; \Theta), \vec{z}^{<t+1, t+\tau_o>}_i), \\
		& = \arg \min_{\Theta} \sum_{i=1}^m \sum_{t=1}^{\tau_o} l_i^{<t>} (Q^{<t>} (\vec{s}^{<t-\tau_i, t>}_i; \Theta), \vec{z}^{<t>}_i),
	\end{split}
	\label{eq:nn_mle_training}
\end{equation}

\noindent where an input sequence of observations $\vec{s}^{<t-\tau_i, t>} = [\vec{s}^{<t-\tau_i>}, ..., \vec{s}^{<t-1>}, \vec{s}^{<t>}]$ is composed of $\tau_i$ consecutive data samples, $l(\cdot,\cdot)$ is the regression loss function and $t$ represents a temporal index. 


Behavioral cloning is a supervised learning approach that directly maps states or observations to actions, given the dataset of expert demonstrations $\mathcal{D}$. Although relatively simple to implement, behavioral cloning strugles with generalization, particularly in situations that differ from the training data. In order to overcome this, in our approach we use Inverse Reinforcement Learning (IRL), which seeks to infer the underlying reward function that the expert is optimizing~\cite{9904958_DRL_Survey, 9537731_inverse_RL}. In the following, we present our proposed IRL approach for training $\vec{h}(\cdot)$.

\section{Scene Dynamics Model Training}
\label{sec:dynamics_model}

We formulate the scene dynamics learning problem in an Inverse Reinforcement Learning (IRL) setup for an autonomous vehicle.

Together with Behavioral Cloning, IRL is a subset of Imitation Learning methods, aiming to learn an expert's behavior from a series of demonstrations.   Essentially, IRL tries to understand why the experts behaves the way they do (i.e. what is the reward they are optimizing for) rather than just copying their actions, thus making the system prone to generalization errors. Once the reward function has been determined, the system learns a policy to maximize the infered reward.

Our goal is to develop a scene dynamics model that, when integrated with the vehicle's nominal process model, can replicate how a human driver maneuvers a vehicle based on its state (position, heading, and velocity) and the evolving driving scene (e.g. movement of other traffic participants, road conditions, etc.).

In our case, the expert is a human driver, for which we store sensory and control data. When acquiring training samples, we store as temporal sequence data: \textit{i}) the historic vehicle states $\vec{z}^{<t-\tau_i, t>}$, \textit{ii}) the range sensing information computed as Occupancy Grids (OG) from fused LiDAR and radar data $\vec{I}^{<t-\tau_i, t>}$, \textit{iii}) the global reference trajectory $\vec{z}^{<t-\tau_i, t+\tau_o>}_{ref}$ and \textit{iv}) the control actions $\vec{u}^{<t-\tau_i, t>}$ recorded from the human driver. A single OG corresponds to an observation instance $\vec{I}^{<t>}$, while a continuous sequence of OGs is denoted as $\vec{I}^{<t-\tau_i, t>}$.

\subsection{Markov Decision Process Modelling}
\label{sec:behavioral_model}

We use the Markov Decision Process (MDP) formalism for modelling the vehicle's behavior in the IRL setup.

The problem can be modeled as a \textit{Markov Decision Process} (MDP) $M := (S, Z_d, \mathcal{T}, R, \gamma)$, where:

\begin{itemize}
	\item $S$ represents a finite set of states, $\vec{s}^{<t>} \in S$ being the joint vehicle and scene state at time $t$. The joint state is defined as the pair $\vec{s}^{<t>} = (\vec{z}^{<t-\tau_i, t>}, \vec{I}^{<t-\tau_i, t>})$.
	
	\item $Z_d$ represents a finite set of vehicle state sequences allowing the vehicle to navigate through the environment observed by $\vec{I}^{<t-\tau_i, t>}$, where $\vec{z}^{<t+1, t+\tau_o>}_d \in Z_d$ is the predicted desired policy that the vehicle should follow in the receding interval $[t+1, t+\tau_o]$. A desired policy $\vec{z}^{<t+1, t+\tau_o>}_d$ is defined as the desired state sequence from Eq.~\ref{eq:behavioral_policy}. These are also used by the NMPC to compute the optimal control actions.
	
	
	\item $\mathcal{T}: S \times Z_d \times S \rightarrow [0, 1]$ is a stochastic transition function, where $\mathcal{T}_{\vec{s}^{<t>}, \vec{z}^{<t+1, t+\tau_o>}_d}^{\vec{s}^{<t+\tau_o>}}$ describes the probability of arriving in state $\vec{s}^{<t+\tau_o>}$, after performing $\vec{z}^{<t+1, t+\tau_o>}_d$ in state $\vec{s}^{<t>}$.
	
	\item $R: S \times Z_d \times S \rightarrow \mathbb{R}$ is a scalar reward function which controls the estimation of $\vec{z}^{<t+1, t+\tau_o>}_d$, where $R_{\vec{s}^{<t>}, \vec{z}^{<t+1, t+\tau_o>}_d}^{\vec{s}^{<t+\tau_o>}} \in \mathbb{R}$. For a state transition $\vec{s}^{<t>} \rightarrow \vec{s}^{<t+\tau_o>}$ at time $t$, we define:
	\begin{equation}
		R_{\vec{s}^{<t>}, \vec{z}^{<t+1, t+\tau_o>}_d}^{\vec{s}^{<t+\tau_o>}} = \vec{w} \cdot  || \vec{z}^{<t+1,t+\tau_o>}_d - \vec{z}^{<t+1,t+\tau_o>}_{ref} ||_2
		\label{eq:reward_function}
	\end{equation}
	where $|| \cdot ||_2$ is the L2 norm and $\vec{w}$ is the weighting vector of the reward, learned from training data via the maximum entropy IRL method from~\cite{ZiebartMBD08}. The reward function is a distance feedback, which is smaller if the desired system's state follows a minimal energy trajectory to the reference state $\vec{z}^{<t+\tau_o>}_{ref}$ and large otherwise.
	
	\item $\gamma$ is the discount factor controlling the importance of future versus immediate rewards.
\end{itemize}

Considering the proposed reward function and an arbitrary set-point trajectory $T = [\vec{z}^{<0>}_d, \vec{z}^{<1>}_d, ..., \vec{z}^{<k>}_d]$ in observation space, at any time $\hat{t} \in [0, 1, ..., k]$, the associated cumulative future discounted reward is defined as:

\begin{equation}
	R^{<\hat{t}>} = \sum^{k}_{t=\hat{t}} \gamma^{<t-\hat{t}>} r^{<t>},
	\label{eq:cumulative_reward}
\end{equation}

\noindent where the immediate reward at time $t$ is given by $r^{<t>}$. In reinforcement learning theory, the statement in Eq.~\ref{eq:cumulative_reward} is known as a finite horizon learning episode of sequence length $k$~\cite{9904958_DRL_Survey, Sutton_RL}.

\subsection{Training}

The objective of the scene dynamics model is to find the desired set-point trajectory that maximizes the associated cumulative future reward. We define the optimal action-value function $Q^*(\cdot, \cdot)$ which estimates the maximal future discounted reward when starting in state $\vec{s}^{<t>}$ and following set-points $\vec{z}^{<t+1, t+\tau_o>}_d$:

\begin{equation}
	Q^* (\vec{s}, \vec{z}_d) = \underset{\pi}{\max} \mathbb{E} \text{ } [R^{<\hat{t}>} | \vec{s}^{<\hat{t}>} = \vec{s}, \text{ } \vec{z}^{<t+1, t+\tau_o>}_d = \vec{z}_d, \text{ } \pi],
	\label{eq:optimal_action_value_function}
\end{equation}

\noindent where $\pi$ is a set-points trajectory (action), which is a probability density function over a set of possible trajectories (actions) that can take place in a given state. The optimal action-value function $Q^*(\cdot, \cdot)$ maps a given state $\vec{s}$ to the optimal set-point trajectory $\vec{z}_d$ of the vehicle in any state:

\begin{equation}
	\forall \vec{s} \in S: \pi^* (\vec{s}) = \underset{\vec{z}_d \in Z_d}{\arg\max} Q^* (\vec{s}, \vec{z}_d).
\end{equation}

The optimal action-value function $Q^*$ satisfies the Bellman optimality equation~\cite{Bellman}, which is a recursive formulation of Eq.~\ref{eq:optimal_action_value_function}:

\begin{equation}
	\begin{split}
		Q^* (\vec{s}, \vec{z}_d) & = \sum_{\vec{s}} \mathcal{T}_{\vec{s}, \vec{z}_d}^{\vec{s}'} \left( R_{\vec{s}, \vec{z}_d}^{\vec{s}'} + \gamma \cdot \underset{\vec{z}'_d}{\max} Q^* (\vec{s}', \vec{z}'_d) \right) \\
		& = \mathbb{E}_{\vec{z}'_d} \left( R_{\vec{s}, \vec{z}_d}^{\vec{s}'} + \gamma \cdot \underset{\vec{z}'_d}{\max} Q^* (\vec{s}', \vec{z}'_d) \right),
	\end{split}
	\label{eq:bellman_optimality_equation}
\end{equation}

\noindent where $\vec{z}_d = \vec{z}^{<t+1, t+\tau_o>}_d$, $\vec{s}' = \vec{s}^{<t+\tau_o>}$ represents a possible state visited after $\vec{s} = \vec{s}^{<t>}$ and $\vec{z}'_d = \vec{z}'^{<t+1, t+\tau_o>}_d$ is the trajectory. The model-based policy iteration algorithm was introduced in~\cite{Sutton_RL}, based on the proof that the Bellman equation is a contraction mapping~\cite{Watkins_Q_Learning} when written as an operator $\nu$:

\begin{equation}
	\forall Q, \lim_{n \rightarrow \infty} \nu^{(n)} (Q) = Q^*.
\end{equation}

However, the standard reinforcement learning method described above, which involves the maintainace of $Q^*$ in a table-like structure, is not feasible due to the high dimensional state space. Instead of the traditional approach, we use a non-linear parametrization of $Q^*$, encoded in the deep neural network illustrated in Fig.~\ref{fig:neural_network_diagram}. In literature, such a non-linear approximator is called a Deep Q-Network (DQN)~\cite{mnih2015humanlevel, 9904958_DRL_Survey} and is used for estimating the approximate action-value function $Q^*$:

\begin{equation}
	Q (\vec{s}^{<t>}, \vec{z}^{<t+1, t+\tau_o>}_d; \Theta) \approx Q^* (\vec{s}^{<t>}, \vec{z}^{<t+1, t+\tau_o>}_d),
\end{equation}

\noindent where $\Theta$ are the parameters of the deep network from Fig.~\ref{fig:neural_network_diagram}.

By taking into account the Bellman optimality equation (Eq.~\ref{eq:bellman_optimality_equation}), it is possible to train a deep network in an inverse reinforcement learning manner through the minimization of the mean squared error. The optimal expected $Q$ value can be estimated within a training iteration $i$ based on a set of reference parameters $\bar{\Theta}_i$ calculated in a previous iteration $i'$:

\begin{equation}
	\vec{z}_d = R_{\vec{s}, \vec{z}_d}^{\vec{s}'} + \gamma \cdot \underset{\vec{z}'_d}{\max} Q(\vec{s}', \vec{z}'_d; \bar{\Theta}_i),
	\label{eq:optimal_q_value}
\end{equation}

\noindent where $\bar{\Theta}_i := \Theta_{i'}$. The new estimated network parameters at training step $i$ are evaluated using the following mean squared error function:

\begin{equation}
	\hat{\Theta}_i = \underset{\Theta_i}{\min} \text{ } \mathbb{E}_{\vec{s}, \vec{z}_d, r, \vec{s}'} \left[ \left( \vec{z}_d - Q(\vec{s}, \vec{z}_d; \Theta_i) \right)^2 \right],
	\label{eq:rl_setup}
\end{equation}

\noindent where $r = R_{\vec{s}, \vec{z}_d}^{\vec{s}'}$. Based on~\ref{eq:rl_setup}, we apply the maximum likelihood estimation function from Eq.~\ref{eq:nn_mle_training} for calculating the weights of the deep network. The gradient is approximated with random samples in the backpropagation through time algorithm, which uses stochastic gradient descent for training:

\begin{equation}
	\nabla_{\Theta_i} = \mathbb{E}_{\vec{s}, \vec{z}_d, r, \vec{s}'} \left[ \left( \vec{z}_d - Q(\vec{s}, \vec{z}_d; \Theta_i) \right) \nabla_{\Theta_i} \left( Q(\vec{s}, \vec{z}_d; \Theta_i) \right) \right].
	\label{eq:gradient_descent}
\end{equation}

At each training phase, the optimization problem~\ref{eq:gradient_descent} was solved using the Adam optimizer for 25 epochs, with a learning rate of $0.001$, an L2 regularization factor $\lambda_2 = 0.0001$ and a batch size of $64$ samples. Dropout was applied to avoid over-fitting.

\begin{algorithm}
    \caption{Inverse Reinforcement Learning training approach for DL-NMPC-SD.}
	\textbf{Input: } Global reference trajectory $\vec{z}^{<t-\infty, t+\infty>}_{ref}$, dataset of training sequences $\mathcal{D}$, initial dynamics model and reward weights $\Theta_0$ and $\vec{w}_0$. \\
 	\textbf{Output:} Learned scene dynamics model weights $\Theta^*$. \\
 	\textit{Reward function learning:}
    	\begin{algorithmic}
		\For {$i \in [0, K_1]$} \\
			\vspace{0.3em}
			\hspace*{\algorithmicindent} Sample $B \leftarrow (\vec{s}^{<t>}, \vec{z}^{<t+1, t+\tau_o>}_d, \vec{s}^{<t+\tau_o>}) \subset \mathcal{D}$ \\
			\vspace{0.3em}
			\hspace*{\algorithmicindent} Estimate $\Theta_i = \arg \max_{\Theta_i} \mathcal{L} (\Theta_i; B) \rightarrow$ eq.~\ref{eq:nn_mle_training} \\
			\vspace{0.3em}
			\hspace*{\algorithmicindent} Compute $R = \vec{w}_i \cdot  || Q(\vec{s}, \Theta_i) - \vec{z}^{<t+1,t+\tau_o>}_{ref} ||_2 \rightarrow$ eq.~\ref{eq:reward_function} \\
			\vspace{0.3em}
			\hspace*{\algorithmicindent} $\nabla_{\vec{w}_i} = \log(Q(\vec{s}, \Theta_i)) \times \vec{z}_{ref}$ \\
			\vspace{0.3em}
			\hspace*{\algorithmicindent} Update $\vec{w}_{i+1} \leftarrow \vec{w}_i - \alpha_1 \nabla_{\vec{w}_i}$
			\vspace{0.3em}
		\EndFor
	\end{algorithmic}
	\vspace{0.2em}
	\textit{Scene dynamics model learning:}
    	\begin{algorithmic}
      	\For {$i \in [0, K_2]$} \\
      		\vspace{0.3em}
			\hspace*{\algorithmicindent} Sample $B \leftarrow (\vec{s}^{<t>}, \vec{z}^{<t+1, t+\tau_o>}_d, \vec{s}^{<t+\tau_o>}) \subset \mathcal{D}$ \\
			\vspace{0.3em}
			\hspace*{\algorithmicindent} Compute $\vec{z}_d = R_{\vec{s}, \vec{z}_d}^{\vec{s}'} + \gamma \cdot \underset{\vec{z}'_d}{\max} Q(\vec{s}', \vec{z}'_d; \Theta_i)$ \\
			\vspace{0.3em}
			\hspace*{\algorithmicindent} $\nabla_{\Theta_i} = \mathbb{E}_{\vec{s}, \vec{z}_d, r, \vec{s}'} \left[ \left( \vec{z}_d - Q(\vec{s}, \vec{z}_d; \Theta_i) \right) \nabla_{\Theta_i} \left( Q(\vec{s}, \vec{z}_d; \Theta_i) \right) \right]$ \\
			\vspace{0.3em}
			\hspace*{\algorithmicindent} Update $\Theta_{i+1} \leftarrow \Theta_i - \alpha_2 \nabla_{\Theta_i}$
			\vspace{0.3em}
		\EndFor \\
      \Return $\Theta^*$
    \end{algorithmic}
    \label{alg:training}
\end{algorithm}

The pseudocode of the IRL training approach is outlined in Algorithm~\ref{alg:training}. The training process consists of two consecutive loops, with each loop sampling a batch $B$ of trajectories from the Augmented Memory, populated with data from the training dataset. In the first loop, the weights $\vec{w}$ of the reward function are learned by training a model in a supervised fashion on batch $B$ (Eq.~\ref{eq:nn_mle_training}). Once a reward value is obtained using Eq.~\ref{eq:reward_function}, we compute the gradient between the reference $\vec{z}_{ref}$ and the output of the model. The weights of the reward function are then updated using gradient descent.

The final weigths of the scene dynamics model are learned in the second loop, where we use the learned reward function to obtain the optimal expected $Q$ value (Eq.~\ref{eq:optimal_q_value}) and the gradient $\nabla_{\Theta}$ (Eq.~\ref{eq:gradient_descent}), which is further used to adapt the model's weights $\Theta$. An ablation study for different configurations of the input convolutional layer is presented in Appendix~B.

\section{Experiments}
\label{sec:experiments}

\begin{figure}
	\centering
	\begin{center}
		\includegraphics[scale=0.8]{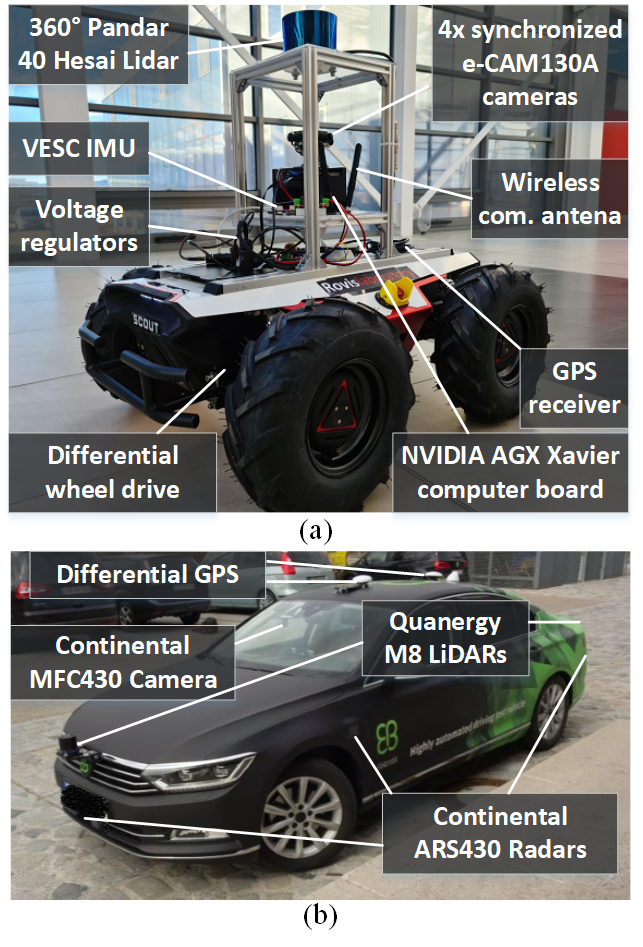}
	    	\vspace{-1em}
		\caption{\textbf{Test vehicles used for data acquisition and testing.} (a) RovisLab AMTU (Autonomous Mobile Test Unit), as a 1:4 scaled car model. (b) Real-sized VW Passat autonomous test vehicle.}
        \label{fig:eb_car}
	\end{center}
    	\vspace{-1.8em}
\end{figure}

The performance of the proposed system was benchmarked in a full closed-loop manner against the classical, learning-free, Dynamic Window Approach (DWA) baseline method~\cite{Fox_Dynamic_Window_Approach_1997}, as well as against two state-of-the-art End2End~\cite{10614862_end2end} and Deep Q-Learning (DQL)~\cite{mnih2015humanlevel, 9904958_DRL_Survey} autonomous driving approaches, respectively.

We have tested DL-NMPC-SD on three different environments: \textit{I}) in the GridSim simulator, \textit{II}) in indoor and outdoor navigation using the RovisLab AMTU (Autonomous Mobile Test Unit) from Fig.~\ref{fig:eb_car}(a) and \textit{III}) on real roads with the full scale autonomous driving car shown in Fig.~\ref{fig:eb_car}(b). This resulted in $40km$ of driving in GridSim, $2km$ of indoor navigation and over $150km$ of driving with DL-NMPC-SD on real-world roads.

The \textit{I} and \textit{II} experiments (Sections~\ref{sec:experiments_A} and ~\ref{sec:experiments_B}) compared DWA, DQL, End2End and the proposed DL-NMPC-SD algorithm. The four variants had to solve the same optimization problem illustrated in Fig.~\ref{fig:problem_description}, which is to calculate a control strategy for safely navigating the driving environment from a starting position to destination. The experiments set \textit{III} (Section~\ref{sec:experiments_C}) tested the DWA and the DL-NMPC-SD algorithms over the course of three trials, each trial being composed of 150km of driving on the highway, inner-city roads and curved country roads, respectively.

In the followings, we describe the evaluation metrics and the experimental results on the three considered benchmarks. The experimental setups, parameters and competing algorithms are presented in Appendices C and E, respectively.

\subsection{Evaluation Metrics}

\begin{figure*}
	\centering
	\begin{center}
		\includegraphics[scale=0.89]{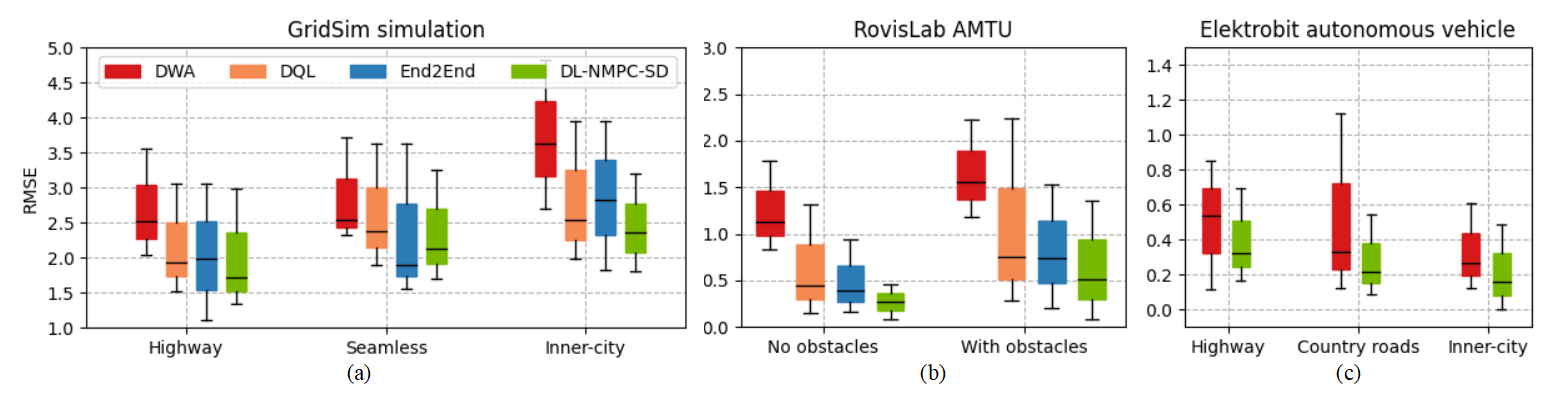}
	    	\vspace{-1.3em}
		\caption{\textbf{Median and variance of RMSE for the three testing environments}. The results show a lower error rate in the case of highway driving, or when the navigation setup does not include obstacles, and higher RMSE for more complex environments, such as inner-city driving.}
        \label{fig:statistical_analysis}
	\end{center}
    	\vspace{-1em}
\end{figure*}

\begin{table*}
	\centering
	\begin{tabular}{ccrlllll}
		\\ \cline{4-8}
		\rule{0pt}{10pt} & & & \multicolumn{3}{l}{\textbf{Quantitative evaluation}} & \multicolumn{2}{l}{\textbf{Accuracy}} \\
		\hline
		\rule{0pt}{10pt} \textbf{Scenario} & \textbf{Description} & \textbf{Method} & \textbf{\% Crashes} & \textbf{\% Reached goal} & \textbf{Avg. Speed [m/s]} & \textbf{$e_L \pm$ STD [m]} & \textbf{$e_H \pm$ STD [deg]} \\
		\hline
		\rule{0pt}{10pt} \multirow{12}{*}{Simulation} & \multirow{4}{*}{Highway} & DWA~\cite{Fox_Dynamic_Window_Approach_1997} & 22.4 & 81.0 & 1.94 & 1.581 & 10.51 \\
		& & DQL~\cite{mnih2015humanlevel} & 31.6 & 63.0 & 4.85 & 1.038 & 19.10 \\
		& & End2End~\cite{10614862_end2end} & 29.5 & 65.0 & 5.29 & 1.171 & 17.83 \\
		& & \textbf{DL-NMPC-SD (ours)} & \textbf{11.8} & \textbf{88.0} & \textbf{6.17} & \textbf{1.009} & \textbf{8.41} \\ \cline{3-8}
		\rule{0pt}{10pt} & \multirow{4}{*}{Seamless} & DWA~\cite{Fox_Dynamic_Window_Approach_1997} & 19.2 & 71.0 & 3.58 & 1.493 & 13.99 \\
		& & DQL~\cite{mnih2015humanlevel} & 32.7 & 61.0 & 4.96 & 1.401 & 18.16 \\
		& & End2End~\cite{10614862_end2end} & \textbf{21.1} & \textbf{73.0} & 5.52 & \textbf{1.115} & \textbf{10.83} \\
		& & \textbf{DL-NMPC-SD (ours)} & 22.7 & 69.0 & \textbf{5.93} & 1.251 & 12.93 \\ \cline{3-8}
		\rule{0pt}{10pt} & \multirow{4}{*}{Inner-city} & DWA~\cite{Fox_Dynamic_Window_Approach_1997} & 25.8 & 81.0 & 2.71 & 1.585 & 17.35 \\
		& & DQL~\cite{mnih2015humanlevel} & 24.0 & 66.0 & 3.87 & 1.493 & 20.91 \\
		& & End2End~\cite{10614862_end2end} & 29.8 & 63.0 & 5.11 & 1.662 & 19.51 \\
		& & \textbf{DL-NMPC-SD (ours)} & \textbf{18.5} & \textbf{83.0} & \textbf{6.54} & \textbf{1.391} & \textbf{13.82} \\ \cline{1-8}
		\hline
	\end{tabular}
	\caption{Simulation results for autonomous vehicle control, where STD represents the standard deviation.}
	\label{tab:results_gridsim}
    	\vspace{-2em}
\end{table*}

As performance metrics, we have evaluated the lateral and heading errors $e_L$ and $e_H$, respectively, as we well as the percentage of times an algorithm crashed the ego-vehicle, the number of times the destination goals were reached and the average speed.

\subsubsection{Lateral error}

Ideally, each method should navigate the environment collision-free, at maximum speed and as close as possible to the ground truth. In order to quantify the total system error, we have chosen the cumulative speed-weighted lateral error from~\cite{CodevillaLKD18} as performance metric:

\begin{equation}
	e_L = \frac{1}{m} \norm{ \sum^{\tau_o}_{t=0} (\hat{\vec{p}}^{<t+i>} - \vec{p}^{<t+i>}) \cdot v_{t+i} }_1,
	\label{eq:rmse}
\end{equation}

\noindent where $\norm{ \cdot }_i$ is the L1 norm and $\hat{\vec{p}}$ and $\vec{p}$ are coordinates on the estimated and ground truth trajectories, respectively. $e_L$ is used to assess how well the ego-vehicle maintains its desired path in terms of lateral positioning relative to the ground truth, calculated as the mean of the driving trajectories acquired during data acquisition.  It reflects the distance between the vehicle’s current position and its intended trajectory, measured perpendicular to the vehicle’s direction of travel. The lateral error $e_L$ is multiplied by the vehicle’s speed at each time step. Errors at higher speeds contribute more significantly to the cumulative metric because maintaining accurate positioning is more critical when the vehicle is moving faster, since high speed deviations can lead to more severe safety risks.

\subsubsection{Heading error}

$e_H$ is also weighted by the speed of the vehicle. It indicates the angular difference between the vehicle’s current heading angle $\hat{\rho}$, oriented in the moving direction, and the heading angle from the ground truth data $\rho$:

\begin{equation}
	e_H = \frac{1}{m} \norm{ \sum^{\tau_o}_{t=0} (\hat{\rho}^{<t+i>} - \rho^{<t+i>}) \cdot v_{t+i} }_1,
	\label{eq:rmse}
\end{equation}

\noindent High heading errors can indicate poor vehicle control, leading to erratic steering behavior, discomfort for passengers, and safety risks.

\subsubsection{\% crashes}

The percentage of times an algorithm causes the ego-vehicle to crash serves as a critical performance metric that directly reflects the safety and reliability of the system. A crash is considered a velocity drop longer than $2s$.

\subsubsection{\% reached goals}

It refers to the ratio of successfully completed routes to the total number of attempts, expressed as a percentage. An attempt is considered successful when the ego vehicle reaches the end of the global reference trajectory $\vec{z}^{<t-\infty, t+\infty>}_{ref}$, without taking into account other factors such as major deviations from the reference.

\subsubsection{Average speed}

It is measured in $m/s$ and defined as the total distance traveled by the ego-vehicle divided by the total time taken to complete the route. The measure helps assess how efficiently the vehicle is completing a route, a higher average speed indicating a better performance.

For statistical analysis, we have normalized and combined the lateral and heading errors as a Root Mean Squared Error (RMSE), resulting in the median and variance of the errors shown in Fig.~\ref{fig:statistical_analysis}. In the following subsections, we will discuss the results of our evaluation in each testing setup, including discussions on the statistical results from Fig.~\ref{fig:statistical_analysis}.

\begin{table*}
	\centering
	\begin{tabular}{ccrlllll}
		\\ \cline{4-8}
		\rule{0pt}{10pt} & & & \multicolumn{3}{l}{\textbf{Quantitative evaluation}} & \multicolumn{2}{l}{\textbf{Accuracy}} \\
		\hline
		\rule{0pt}{10pt} \textbf{Scenario} & \textbf{Description} & \textbf{Method} & \textbf{\% Crashes} & \textbf{\% Reached goal} & \textbf{Avg. Speed [m/s]} & \textbf{$e_L \pm$ STD [m]} & \textbf{$e_H \pm$ STD [deg]} \\
		\hline
		\rule{0pt}{10pt} \multirow{8}{*}{Indoor} & No obstacles & DWA~\cite{Fox_Dynamic_Window_Approach_1997} & 21.2 & 73.0 & 0.83 & 1.142 & 8.00 \\
		& on the & DQL~\cite{mnih2015humanlevel} & 30.2 & 64.0 & 1.29 & 0.204 & 18.31 \\
		& ground & End2End~\cite{10614862_end2end} & 29.0 & 61.0 & \textbf{1.66} & 0.129 & 18.31 \\
		& truth path & \textbf{DL-NMPC-SD (ours)} & \textbf{13.8} & \textbf{85.0} & 1.39 & \textbf{0.063} & \textbf{4.51} \\ \cline{3-8}
		\rule{0pt}{10pt} & Obstacles & DWA~\cite{Fox_Dynamic_Window_Approach_1997} & 39.2 & 49.0 & 1.17 & 1.296 & 11.49 \\
		& present on & DQL~\cite{mnih2015humanlevel} & 41.0 & 47.0 & 0.92 & 0.391 & 24.85 \\
		& the ground & End2End~\cite{10614862_end2end} & 45.0 & 43.0 & 1.48 & 0.337 & 19.38 \\
		& truth path & \textbf{DL-NMPC-SD (ours)} & \textbf{23.1} & \textbf{76.0} & \textbf{1.59} & \textbf{0.103} & \textbf{7.94} \\ \cline{1-8}
		\hline
	\end{tabular}
	\caption{Autonomous control results for the RovisLab AMTU robot in Fig.~\ref{fig:eb_car}(a), where STD represents the standard deviation.}
	\label{tab:results_model_car}
    	\vspace{-1.5em}
\end{table*}

\subsection{Experiment I: Simulation Results}
\label{sec:experiments_A}

For the three GridSim scenarios, quantitative and qualitative performance evaluation results are shown in Table~\ref{tab:results_gridsim}, as well as in the statistical analysis of the RMSE from Fig.~\ref{fig:statistical_analysis}(a).

The first set of experiments compared the four algorithms (DWA, End2End, DQL and DL-NMPC-SD) on over 10 goal-navigation trials in the GridSim~\cite{Trasnea_IRC2019} simulation environment, configured using the parameters from Appendix~C. The task was to reach a given goal location from a starting position, while avoiding collisions and driving at maximum speed.

\textit{Training}: To train the competing End2End and DQL methods, we conducted the goal navigation task on 10 different driving routes. In the case of End2End learning, we have built a trajectories database mapped to sensory information, while the ego-vehicle was manually driven by a person. This resulted in over $50.000$ pairs of OG sensory samples mapped to human steering commands, over the $10$ driving routes. The collected database was used for training the End2End learning system. In the DQL case, the ego-vehicle was trained in an episodic manner, by exploring the driving environment defined by the $10$ trials, according to the setup described in Appendix~C.

\textit{Results and discussion}: In simulation scenarios involving highway driving, our method achieved the lowest percentage of crashes (11.8\%) and the highest number of reached destination goals (88\%), while maintaining an average speed of $6.17m/s$, which is the highest among the four tested methods. In the case of the seamless environment, the End2End approach outperformed the other methods. This is however expected, since the seamless environment does not contain any structure, just roads which are generating data that can be overfitted by the End2End model. The case of inner-city driving in simulation is actually the closest scenario to real-world driving. It encompases U-turns, T-junctions, roundabouts or variable lanes. DL-NMPC-SD outperforms the other approaches in the considered performance metrics.

In all simulation experiments, except the seamless environment, DL-NMPC-SD exhibits the lowest lateral and heading errors, despite the fact that DWA was exhaustively searching for the best trajectory route, given the obstacles present in the scene.

In comparison to the other two learning based methods (DQL and End2End), our approach yielded a significantly lower heading error in highway and inner-city driving. Given that neither DQL nor End2End relies on a nominal vehicle model, the higher heading error is likely attributed to their exploratory nature, as they do not take into account the vehicle's kinematic constraints.

In terms of training data, we have epirically observed that the accuracy of the learning systems is proportional to the variance and amount of training data used. In this specific setup, the training data and trajectory of the traffic participants was similar to the given training information.

One of the biggest challenges when using data-driven techniques for control is the so-called "DaGGer effect"~\cite{Joonwoo_Dagger_2024}, which is a significant drop in performance when the training and testing trajectories are significantly different. The DaGGer effect~\cite{Joonwoo_Dagger_2024} could be avoided, since the simulation environment allowed us to gather as much training data as necessary.

In terms of percentage of reached goals, the DWA was the only method coming closer to the DL-NMPC-SD results, but with the expense of the average travel speed.

\subsection{Experiment II: Indoor and Outdoor Testing Results}
\label{sec:experiments_B}

The second experiment compared the four algorithms using the real world RovisLab AMTU vehicle from Fig.~\ref{fig:eb_car}(a), configured using the parameters described in Appendix~C. The task was to safely navigate two indoor and outdoor loops, each having a $120m$ length. The quantitative and qualitative performance evaluation results are shown in Table~\ref{tab:results_model_car} and Fig.~\ref{fig:statistical_analysis}(b), while the maximum tracking errors and travel time vs. trial number are shown in Appendix~F.


\textit{Training}: For gathering training data, we have driven AMTU on indoor and outdoor test tracks, gathering $42.000$ samples. The OGs are computed using the pipeline described in Appendix~D, yielding perception data similar to the one calculated within the VW Passat test car from Fig.~\ref{fig:eb_car}(b). The methods have been trained on the whole dataset, since the testing took place online and did not rely on a test dataset. Once trained, the algorithms had to solve the goal navigation task in the same driving environments for $10$ different trials. The mean of the driving trajectories acquired during training data acquisition is considered as the ground truth path. 

\textit{Results and discussion}: As shown in Table~\ref{tab:results_model_car}, DL-NMPC-SD outperformed the other methods on all performance metrices, except on the average speed in the case of autonomous navigation on a trajectory with no obstacles. Although the experiments were conducted in both indoor and outdoor settings, we did not observe any significant statistical difference in the errors between the two setups, suggesting that the observations are independent of the environment.

In the ``No obstacles on the ground truth path'' scenario, DL-NMPC-SD outperformed the other methods, showing the lowest crash rate (13.8\%) and the highest goal-reaching rate (85.0\%). Additionally, it achieves the best accuracy with the lowest lateral error ($e_L=0.063m$) and heading error ($e_H=4.51^{\circ}$), while maintaining a competitive speed ($1.39 m/s$). Only End2End achieves a higher speed ($1.66 m/s$) but at the cost of higher crash rates and lower accuracy.

In the ``Obstacles on the ground truth path'' scenario, DL-NMPC-SD again shows superior performance, reducing the crash rate to 23.1\%, while achieving a 76.0\% goal-reaching rate, the highest among the methods. It also maintains low lateral ($e_L=0.103 m$) and heading ($e_H=7.94^{\circ}$) errors, indicating better precision in navigation compared to DWA, DQL, and End2End.

In order to assess the behavior of the methods with respect to the DaGGer effect, we have placed on the reference route obstacles which were not given at training time. The DQL and End2End results have the largest heading tracking error. In comparison, DL-NMPC-SD computes a smoother trajectory, yielding lower path tracking errors.

The accuracy of DL-NMPC-SD is especially visible in the experiments containing obstacles, where we have obtained a lower percentage of crashes, as well as a higher number of reached goals. One reason for the DQL and End2End's performance decrease is the discrete nature of the steering commands provided as control input by the two approaches, which is particuraly present in the real-world test setups.
The effect of the discrete steering commands calculated by the DQL and End2End controllers are visible in the tracking errors as a jittering phenomenon. The jittering refers to rapid, small-scale oscillations or fluctuations in the vehicle's control decisions, manifesting in steering and velocity adjustments. This confirmed our objective to design a learning based scene dynamics NMPC that can predict and adapt the trajectory of an ego-vehicle smoothly, according to the dynamics of the driving environment.

\subsection{Experiment III: On-Road Testing Results}
\label{sec:experiments_C}

\begin{table*}
	\centering
	\begin{tabular}{ccrlllll}
		\\ \cline{4-8}
		\rule{0pt}{10pt} & & & \multicolumn{3}{l}{\textbf{Accuracy}} & \\
		\hline
		\rule{0pt}{10pt} \textbf{Scenario} & \textbf{Description} & \textbf{Method} & \textbf{Max $e_L$ [m]} & \textbf{Max $e_H$ [deg]} & \textbf{Avg. Speed [m/s]} & \textbf{$e_L \pm$ STD [m]} & \textbf{$e_H \pm$ STD [deg]} \\
		\hline
		\rule{0pt}{10pt} \multirow{4}{*}{Highway} & \multirow{2}{*}{Highway trial 1} & DWA~\cite{Fox_Dynamic_Window_Approach_1997} & 0.180 & 17.32 & 25 & 0.037 & 2.90 \\
		& & \textbf{DL-NMPC-SD (ours)} & \textbf{0.103} & \textbf{9.04} & 25 & \textbf{0.026} & \textbf{1.46} \\ \cline{3-8}
		\rule{0pt}{10pt} & \multirow{2}{*}{Highway trial 2} & DWA~\cite{Fox_Dynamic_Window_Approach_1997} & 0.201 & 20.91 & 27.77 & 0.115 & 4.18 \\
		& & \textbf{DL-NMPC-SD (ours)} & \textbf{0.119} & \textbf{11.86} & 27.77 & \textbf{0.099} & \textbf{2.95} \\ \cline{1-8}
		\rule{0pt}{10pt} \multirow{4}{*}{Country roads} & \multirow{2}{*}{Country roads 1} & DWA~\cite{Fox_Dynamic_Window_Approach_1997} & \textbf{0.245} & \textbf{14.28} & 18.05 & \textbf{0.093} & \textbf{4.69} \\
		& & \textbf{DL-NMPC-SD (ours)} & 0.285 & 18.15 & 18.05 & 0.105 & 5.01 \\ \cline{3-8}
		\rule{0pt}{10pt} & \multirow{2}{*}{Country roads 2} & DWA~\cite{Fox_Dynamic_Window_Approach_1997} & 0.241 & 23.85 & 17.22 & 0.194 & 4.00 \\
		& & \textbf{DL-NMPC-SD (ours)} & \textbf{0.229} & \textbf{16.93} & 17.22 & \textbf{0.127} & \textbf{3.86} \\ \cline{1-8}
		\rule{0pt}{10pt} \multirow{4}{*}{Inner-city} & \multirow{2}{*}{Inner-city roads 1} & DWA~\cite{Fox_Dynamic_Window_Approach_1997} & 0.252 & 25.39 & 13.88 & 0.083 & 5.40 \\
		& & \textbf{DL-NMPC-SD (ours)} & \textbf{0.152} & \textbf{9.68} & 13.88 & \textbf{0.047} & \textbf{2.68} \\ \cline{3-8}
		\rule{0pt}{10pt} & \multirow{2}{*}{Inner-city roads 2} & DWA~\cite{Fox_Dynamic_Window_Approach_1997} & 0.407 & 28.05 & 13.05 & 0.158 & 6.51 \\
		& & \textbf{DL-NMPC-SD (ours)} & \textbf{0.207} & \textbf{13.59} & 13.05 & \textbf{0.094} & \textbf{3.84} \\
		\hline
	\end{tabular}
	\caption{Autonomous control results for the real size car in Fig.~\ref{fig:eb_car}(b).}
	\label{tab:results_eb_car}
    \vspace{-2em}
\end{table*}

The third experiment tested the DL-NMPC-SD algorithm on over $150km$ of driving in highway, country roads and inner-city environments, according to the setup from Appendix~C. Due to the jittering effect encountered in the DQL and End2End controllers, we have decided to benchmark DL-NMPC-SD only against DWA in real-world driving conditions, with the test car from Fig.~\ref{fig:eb_car}(b). Accuracy results are given in Table~\ref{tab:results_eb_car}, as well as in the statistical analysis from Fig.~\ref{fig:statistical_analysis}(c). Path tracking errors are detailed in Appendix~F.

\textit{Training}: We have used the same training procedure as in the case of the RovisLab AMTU experiments described in the previous section. However, in this scenario, $75\%$ of the $463.000$ collected samples have been used for training, while $15\%$ have been used for validation. The remaining $15\%$ have been divided into two testing trial batches per scenario.

\textit{Results and discussion}: DL-NMPC-SD outperforms DWA in all trials, with lower lateral and heading errors. For instance, in the highway trial 1, DL-NMPC-SD achieves a maximum lateral error of $0.103 m$ compared to DWA's $0.180 m$, while its heading error is also smaller ($9.04^{\circ}$ vs. $17.32^{\circ}$). As shown in the analysis from Fig.~\ref{fig:statistical_analysis}(c), the standard deviations for DL-NMPC-SD are also lower, indicating a smoother and more stable control. The average speed is kept constant for both methods, meaning the improvements come from a better control precision.

DL-NMPC-SD consistently outperforms DWA across all scenarios in terms of lower maximum lateral and heading errors. The errors are expressed with respect to the ground truth calculated from the steering commands of the human driver. DL-NMPC-SD can reliably predict the dynamics of the driving scene by adapting the desired trajectory of the car. This is visible also in the path tracking errors from Appendix~F, illustrated for highway and inner-city driving, respectively. Both algorithms show relatively stable lateral and heading errors, with DL-B-NMPC consistently maintaining a lower error level. The errors in inner-city for both algorithms are more dynamic, reflecting the increased complexity of the urban environment. Nevertheless, DL-B-NMPC exhibits smaller and less frequent error oscillations.

We believe that one of the main reasons for a better accuracy of DL-NMPC-SD is the incorporated reference trajectory $\vec{z}^{<t-\infty, t+\infty>}_{ref}$ in the set-points calculations, making it straightforward to calculate optimal future trajectory states, even at intersection points.

\section{Discussion and Limitations}
\label{sec:discussion}

In this work, we focus solely on the dynamics of the scene observed through sequences of occupancy grids, primarily modeling the interactions between vehicles. Factors such as environmental conditions or traffic rules are not considered, specifically to clearly analyze how combining a nominal model with a scene dynamics model would impact the performance of a constrained NMPC controller. Using the formalism presented in this paper, environmental conditions and traffic rules can be incorporated into the dynamics model by employing additional sensors that detect these elements (e.g. traffic sign and traffic light detectors, lane boundaries, etc.), along with a more sophisticated representation of the reward function.

One of the main practical limitations of DL-NMPC-SD, when deployed in real-world self-driving cars, is the necessity to comply with the functional safety standard ISO26262 and Automotive Safety Integrity Level (ASIL) standards. These standards are crucial for ensuring that autonomous driving systems operate reliably and safely, even in the presence of hardware or software failures. They require systems to be predictable, explainable, and verifiable, which contrasts with the inherent nature of our deep learning based scene dynamics model, which operates as a black box with complex, non-transparent decision-making processes. Although progress has been made in areas such as explainable AI and fault-tolerant architectures, current deep learning implementations do not meet the safety and reliability standards required for ASIL certification. A solution to these problems would be the addition of safety mechanisms, such as redundancy and fallback systems.

Another limitation of the proposed approach is the amount and variance of the training data. Although we have collected an amount of data sufficient for evaluation with respect to state-of-the-art methods, a deployment of DL-NMPC-SD in real-world applications would require the acquisition of data which would contain a larger amount of corner cases and situations. Such a comprehensive dataset would guard against an unstable behaviour in predicting future vehicle states.

A large comprehensive dataset would also guard against the Dagger effect~\cite{Joonwoo_Dagger_2024}, by increasing the generalization capabilities of the neural network. The "DaGGer effect" is also the main reason why an IRL approach, which uses the reward function to explore different trajectories, is preferred over standard supervised behavioral cloning. In the experimental Section~\ref{sec:experiments_B}, we have shown that a high generalization can be achieved, demonstrating that DL-NMPC-SD can safely navigate the driving environment, even if the encountered scene dynamic was not given at training time.

One of the limitations of our method is the linear reward model from~\cite{ZiebartMBD08}, used mainly due to its simplicity. More advanced modeling approaches, such as reward function representations using deep neural networks~\cite{WulfmeierWP16}, have been developed and can be considered for future implementations of DL-NMPC-SD.

When training the MDP using Algorithm~\ref{alg:training}, we have observed changes in the training and validation errors when increasing the number of layers in the convolutional blocks, and/or increasing the number of hidden neurons in the LSTMs. In particular, the training error decreases with additional layers, since the model becomes more capable of fitting the training data. The validation error also decreases, until the model overfits the training data. This happens when the model starts to capture noise in the training data, resulting in poor generalization to new, unseen data. In order to address this issue, we have empirically determined the number of layers and units from Fig.~\ref{fig:neural_network_diagram}, which offer a tradeoff between low training and validation errors, as well as a proper generalization on new unseen data and a low computation time.

\section{Conclusions}
\label{sec:conclusions}

In this work, we have presented a novel method, coined \textit{Deep Learning-based Nonlinear Model Predictive Controller with Scene Dynamics} (DL-NMPC-SD), for controlling autonomous vehicles in different driving conditions. The controller makes use of an a-priori process model and behavioral and disturbance models encoded in the layers of a deep neural network. We show that by incorporating the dynamics of the scene in a deep network, DL-NMPC-SD can be used to steer an autonomous vehicle in different operating conditions, such as indoor and outdoor environments, as well as for autonomous driving on highway, country and inner-city roads. We train our system in an Inverse Reinforcement Learning fashion, based on the Bellman optimality principle.

As future work, the authors plan to investigate the stability of DL-NMPC-SD, especially in relation to the functional safety requirements needed for deployment on a larger scale and through the inclusion of a terminal constraint term in the cost function of the motion controller from Appendix~A. Through high robustness, the proposed method might represent an important component of next-generation autonomous driving technologies, contributing to safer and better self-driving cars.





%

\ifCLASSOPTIONcaptionsoff
  \newpage
\fi



%

\bibliographystyle{IEEEtran}
\bibliography{references}

%

\vspace{-3em}

\begin{IEEEbiography}[{\includegraphics[width=1in,height=1.25in,clip,keepaspectratio]{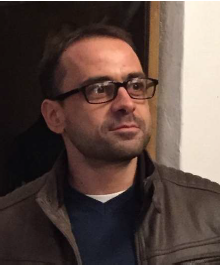}}]{Sorin M. Grigorescu} received the Ph.D. degree in Robotics from the University of Bremen, Germany, in 2010 and the Dipl.-Eng. Degree in Control Engineering and Computer Science from Transilvania University of Brasov, Romania, in 2006. Between 2006 and 2010 he was a member of the Institute of Automation, University of Bremen, where he coordinated the FRIEND service robotics project. Sorin is a professor at the Department of Automation and Information Technology, Transilvania University of Brasov, Romania, where he leads the Robotics, Vision and Control Laboratory (RovisLav) since June 2010. Sorin was an exchange researcher at several institutes, such as the Korea Advanced Institute of Science and Technology (KAIST), the Intelligent Autonomous Systems Group at the Technical University Munich, or the Robotic Intelligence Lab at University Jaume I in Spain His research interests are Artificial Intelligence in Robotics, Simultaneous Localization and Mapping and Learning Controllers.\end{IEEEbiography}

\vspace{-3em}

\begin{IEEEbiography}[{\includegraphics[width=1in,height=1.25in,clip,keepaspectratio]{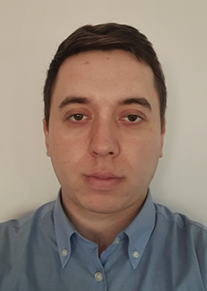}}]{Mihai V. Zaha} received the BE in Control Engineering and Computer Science from the Transilvania University of Brasov, Romania, in 2019, and the MEng degree in System Engineering from the Transilvania University of Brasov, Romania, in 2021. Since oct. 2021 he is affiliated with the Dep. of Automation and Information Technology at Transilvania University of Brasov, where he teaches system and control theory. Since March 2016 he is also affiliated with Elektrobit Automotive Romania, where he develops automotive software. His research interests are Learning Control, State Estimation and Autonomous Robotics.\end{IEEEbiography}







\section*{Appendix A}
\section*{Motion Control using a Scene Dynamics based Constrained NMPC}
\label{sec:motion_control}

Once its weights have been obtained using Algorithm~\ref{alg:training}, the scene dynamics model can be used to improve the prediction capabilities of the contrained NMPC controller.

Traditional NMPC controllers rely on an internal nominal vehicle model to predict the vehicle's future states, allowing the computation of optimal control inputs at each sampling time. To enhance the controller's performance by incorporating the driving scene, we combine the nominal model with the trained scene dynamics model using Eq.~\ref{eq:true_system_model}.

We distinguish between the global reference trajectory $\vec{z}^{<t-\infty, t+\infty>}_{ref}$, which is practically infinite from a control perspective, and a desired set-points trajectory $\vec{z}^{<t+1, t+\tau_o>}_d$, calculated over receding horizon $\tau_o$. By requiring as reference only a global route, we manage to estimate in a closed-loop manner the desired response of the controller. The optimal future states are computed by a constrained NMPC controller, based on a desired state trajectory $\vec{z}^{<t+1, t+\tau_o>}_d$ estimated by the scene dynamics model. When computing the desired set-points trajectory $\vec{z}^{<t+1, t+\tau_o>}_d$, we use the curent control actions $\vec{u}^{<t>}$ as input to all the Process Model components from Fig.~\ref{fig:neural_network_diagram}.

On top of the combined nominal vehicle and scene dynamics model prediction from Eq.~\ref{eq:true_system_model}, we define the cost function to be optimized by the constrained NMPC in receding interval $[t+1, t+\tau_o]$ as:

\begin{equation}
	J (\vec{z}, \vec{u}) = (\vec{z} - \vec{z}_d)^T \vec{Q} (\vec{z} - \vec{z}_d) + \vec{u}^T \vec{R} \vec{u},
	\label{eq:nmpc_cost_function}
\end{equation}

\noindent where $\vec{Q} \in \mathbb{R}^{\tau_o n \times \tau_o n}$ is positive semi-definite, $\vec{R} \in \mathbb{R}^{\tau_o M \times \tau_o M}$ is positive definite and $\vec{u}^{<t, t+\tau_o>} = [\vec{u}^{<t>}, ..., \vec{u}^{<t+\tau_o>}]$ is the control input sequence.

Eq.~\ref{eq:nmpc_cost_function} follows the standard definition of a typical cost function used in model predictive control. $J(\cdot)$ quantifies the performance of the control system over a prediction horizon, in order to determine the optimal control action. In our work, we penalize deviations from desired states and control effort using weighting matrices $\vec{Q}$ and $\vec{R}$, respectively, while omitting the terminal cost. Given a control input and a receding horizon, Eq.~\ref{eq:nmpc_cost_function} quantifies the performance of the control system, in terms of reference state tracking and control effort, that is, the amplitude of the control input. By adjusting $\vec{Q}$ and $\vec{R}$, one can fine-tune the controller to either accurately follow the reference regardless of the control input's amplitude or to prioritize minimizing control effort over tracking accuracy.

The NMPC objective is to find a set of control actions which optimizes the vehicle's trajectory over a given receding horizon $\tau_o$, while satisfying a set of hard and soft constraints:

\begin{subequations}
	\begin{equation}
		(\vec{z}^{<t+1>}_{opt}, \vec{u}^{<t+1>}_{opt}) = \underset{\vec{z}, \vec{u}}{\arg\min} \text{ } J \text{ } (\vec{z}^{<t+1, t+\tau_o>}, \vec{u}^{<t+1, t+\tau_o>})
		\label{eq:nmpc_optimization}
	\end{equation}
	\begin{flalign}
		& \text{such that } \vec{z}^{<0>} = \vec{z}^{<t>} &
	\end{flalign}
	\begin{flalign}
		&
		\begin{split}
			\vec{z}^{<t+i+1>} = & \vec{f} (\vec{z}^{<t>}, \vec{u}^{<t>}) + \vec{h} (\vec{s}^{<t>}),
		\end{split}
		&&
	\end{flalign}
	\begin{flalign}
		&
		\begin{split}
			\vec{e}^{<t+i>}_{\min} \leq \vec{e}^{<t+i>} \leq \vec{e}^{<t+i>}_{\max}, 
		\end{split}
		&&
	\end{flalign}
	\begin{flalign}
		&
		\begin{split}
			\vec{u}^{<t+i>}_{\min} \leq \vec{u}^{<t+i>} \leq \vec{u}^{<t+i>}_{\max}, 
		\end{split}
		&&
	\end{flalign}
	\begin{flalign}
		&
		\begin{split}
			\vec{\dot{u}}^{<t+i>}_{\min} \leq \frac{\vec{\dot{u}}^{<t+i>} - \vec{\dot{u}}^{<t+i-1>}}{\Delta t} \leq \vec{\dot{u}}^{<t+i>}_{\max}, 
		\end{split}
		&&
	\end{flalign}
\end{subequations}

\noindent where $i = 0, 1, ..., \tau_o$, $\vec{z}^{<0>}$ is the initial state and $\Delta t$ is the sampling time. $\vec{e}^{<t+i>} = \vec{z}^{t+i}_d - \vec{z}^{t+i}$ is the cross-track error, while $\vec{e}^{<t+i>}_{min}$ and $\vec{e}^{<t+i>}_{max}$ are the lower and upper tracking bounds, respectively. Additionally, we consider $\vec{u}^{<t+i>}_{\min}$, $\vec{\dot{u}}^{<t+i>}_{\min}$ and $\vec{u}^{<t+i>}_{\max}$, $\vec{\dot{u}}^{<t+i>}_{\max}$ as lower and upper constraint bounds for the actuator and actuator's rate of change, respectively. The DL-NMPC-SD controller implements:

\begin{equation}
	\vec{u}^{<t>} = \vec{u}^{<t+1>}_{opt},
\end{equation}

\noindent at each iteration $t$.

The current optimization procedure from Eq.~\ref{eq:nmpc_optimization} does not include a terminal cost or a terminal constraint on the vehicle state. Such terms are used in order to guaratee a safe trajectory, even for states beyond the considered receding horizon $[t+1, t+\tau_o]$. This becomes particularly problematic when the receding horizon is short. For example, in autonomous driving, a predictive controller that generates a vehicle’s trajectory over a short horizon without considering an approaching curve may accelerate excessively, making it impossible to navigate the turn safely. Since our work focuses on developing an improved process model for predicting future vehicle states within the optimization loop from Eq.~\ref{eq:nmpc_optimization}, rather than enhancing the NMPC algorithm itself, we have opted to omit the terminal constraint term. Furthermore, we use a sufficiently large receding horizon to ensure that the absence of a terminal constraint term does not compromise the system's stability. However, the stability of the closed-loop system can only be guaranteed through an additional terminal cost function term, which is a control Lyapunov function over the terminal set of states. The inclusion of such a term will be considered in future work, where we intend to analyze the overall stability of the system.

We leverage on the quadratic cost function~\ref{eq:nmpc_cost_function} and solve the nonlinear optimization problem described above using the Broyden-Fletcher-Goldfarb-Shanno algorithm~\cite{Flet87}. The quadratic form allows us to apply the quasi-Newton optimization method, without the need to specify the Hessian matrix.

\section*{Appendix B}
\section*{Ablation Study}

\begin{figure*}
	\centering
	\begin{center}
		\includegraphics[scale=0.75]{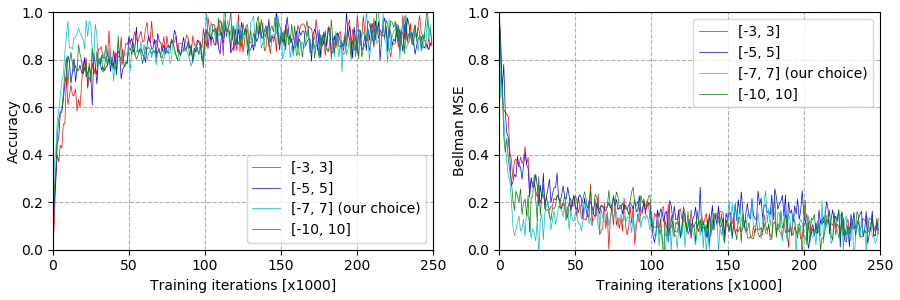}
		\vspace{-1em}
		\caption{\textbf{Training evolution}. Visualization of the accuracy and mean squared error on the Bellman equation for different configurations of the input convolutional layer in Fig.~\ref{fig:neural_network_diagram}.}
	\label{fig:training_evaluation}
      	\vspace{-0.5em}
	\end{center}
\end{figure*}

In order to acquire training data, we have performed data acquisition both in GridSim~\cite{Trasnea_IRC2019}, as well as in real-world navigation and driving scenarios, as detail in the experimental Section~\ref{sec:experiments}. 
While training, we have performed an ablation study on the architecture from Fig.~\ref{fig:neural_network_diagram}, varying the number of convolution filters, the size of the fully-connected layers and the number of hidden LSTM neurons. We empirically observed that the topology illustrated in Fig.~\ref{fig:neural_network_diagram} yields optimal results, both in terms of speed and accuracy. The training criterion was the Bellman error, which measures the quality of the desired states trajectory. Fig.~\ref{fig:training_evaluation} shows the evolution of the accuracy and of the Bellman error during training, for different sizes of the input convolutional layer.


The training procedure was developed in Python, using TensorFlow 2.2. For training, we have used a desktop computer equipped with an Intel Core i7 7700K CPU, 64GB RAM and two high-performance NVIDIA GeForce GTX 1080 Ti graphic cards. The inference procedure was written in C++ and computed on an NVIDIA AGX Xavier board equipped with a 512-core Volta GPU, a 64 bit 8-core ARM CPU and 32GB of RAM.

\section*{Appendix C}
\section*{Testing Scenarios}
\label{sec:testing_scenarios}

\subsection{Simulation Testing Setup in GridSim}
\label{sec:setup_testing_sim}

\begin{figure}
	\centering
	\begin{center}
		\includegraphics[scale=0.85]{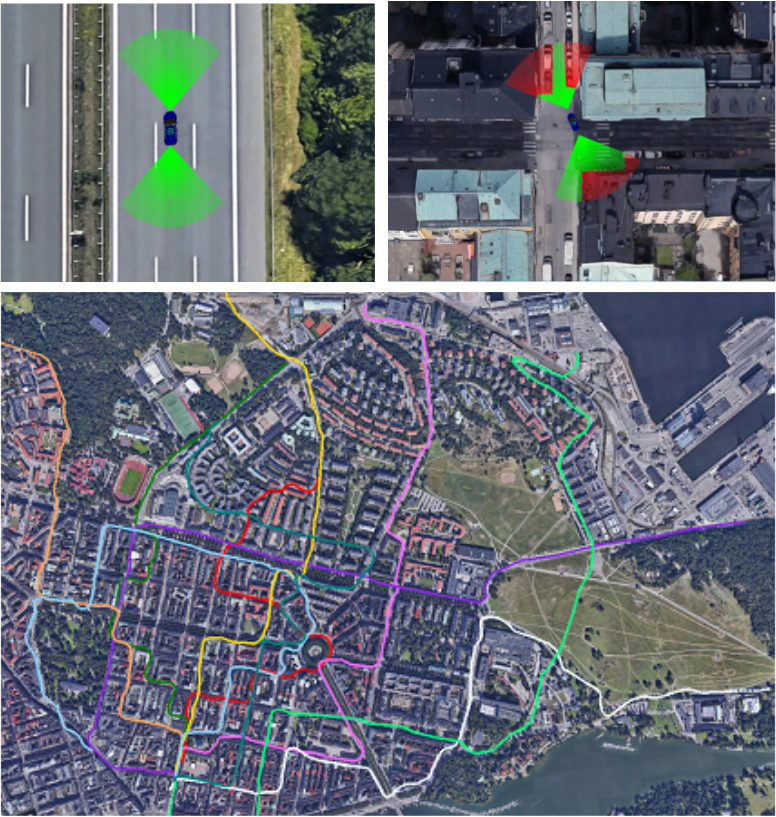}
		\caption{\textbf{Evaluation routes in GridSim~\cite{Trasnea_IRC2019}}. (top) Top-view snapshots during highway and inner-city driving. (bottom) A GridSim virtual test field, defined on $30 km^2$ of the Stockholm inner-city area.}
        \label{fig:stockholm_routes}
	\end{center}
\end{figure}

For simulation testing and synthetic data generation, we utilize our proprietary occupancy grids simulator, GridSim~\cite{Trasnea_IRC2019}. GridSim uses kinematic car models and an OG sensor model to generate synthetic training data, together with their corresponding desired state trajectory labels. Example snapshots from GridSim are shown in Fig~\ref{fig:stockholm_routes}.

GridSim can be used in three simulated environments: seamless with dynamics obstacles only, inner-city and highway driving. The simulation parameters for each environment are given in Table~\ref{tab:simulatoin_parameters}. The sensors have a Field of View (FoV) of $120^{\circ}$. They react when an obstacle is sensed by marking it as an occupied area, according to the range sensing perception algorithm described in Section~\ref{sec:perception}. Examples of simulated OGs are illustrated in Fig.~\ref{fig:occupancy_grids}(a,b).

\begin{table}
	\centering
	\begin{tabular}{llll}
		\hline
		\textbf{Description} & \textbf{Seamless} & \textbf{Inner-city}& \textbf{Highway} \\
		\hline
		Field of View (FoV) & $120^{\circ}$ & $120^{\circ}$ & $120^{\circ}$ \\
		Number of traffic participants & 10 & 20 & 7 \\
		Maximum speed & $13.88 m/s$ & $8.33 m/s$ & $27.77 m/s$ \\
		Maximum acceleration & $2 m/s^2$ & $2 m/s^2$ & $4 m/s^2$ \\
		Minimum speed & $4.16 m/s$ & $2.77 m/s$ & $8.33 m/s$ \\
		Percentage of straight road & $60\%$ & $45\%$ & $100\%$ \\
		Mean curve radius & $55^{\circ}$ & $81^{\circ}$ & $0^{\circ}$ \\
		\hline
	\end{tabular}
	\caption{GridSim simulation parameters.}
	\label{tab:simulatoin_parameters}
    \vspace{-1.5em}
\end{table}

The evaluation routes, illustrated in Fig.~\ref{fig:stockholm_routes}, where defined within a $30 km^2$ inner-city area, mapped in GridSim from the downtown map of Stockholm. Alongside static objects, such as buildings or traffic lights, we have also added dynamic obstacles with which the ego-vehicle has to cope by overtaking.

The dynamic obstacles are represented by other cars, which have their trajectory sampled from a database of pre-recorded driving trajectories. The static obstacles are a-priori mapped to the background as lists of polygons. The simulated sensor continuously checks if the perception rays are colliding with the given polygons. The number of traffic participants in GridSim is configurable. In our evaluation, we have chosen 10, 20 and 7 traffic participants for the highway, inner-city and seamless scenarios, respectively. At each time, the traffic participants interact with the ego-vehicle by overtaking, lane change, following, or braking.

Depending on the simulation scenarios, the maximum speed and acceleration of the traffic participants vary. The maximum speed has a value of $13.88 m/s$, $8.33 m/s$ and $27.77 m/s$ in the three scenarios, respectively, while the maximum acceleration is $2 m/s^2$ for seamless and inner-city driving and $4 m/s^2$ for highway driving. In order not to disturb the traffic, we also impose a minimum speed of $4.16 m/s$, $2.77 m/s$ and $8.33 m/s$ in the three scenarios, respectively.

The structure of the road differs depending on the testing scenario. The percentage of straight road is $60\%$ for seamless driving and $45\%$ for inner-city driving, each having a mean curve radius of $55^{\circ}$ and $81^{\circ}$, respectively. Although curved sections are also present in highway scenarios, we consider them negligible. Therefore, such a scenario is defined by $100\%$ straight roads with $0^{\circ}$ mean curve radius.

Synthetic data is generated by calculating centered OGs at fixed timestamps. At each timestamp, the OGs are saved together with the steering angle and longitudinal velocity.

\subsection{Indoor and Outdoor Testing Setup using RovisLab AMTU}
\label{sec:amtu}

\begin{figure}
	\centering
	\begin{center}
		\includegraphics[scale=0.95]{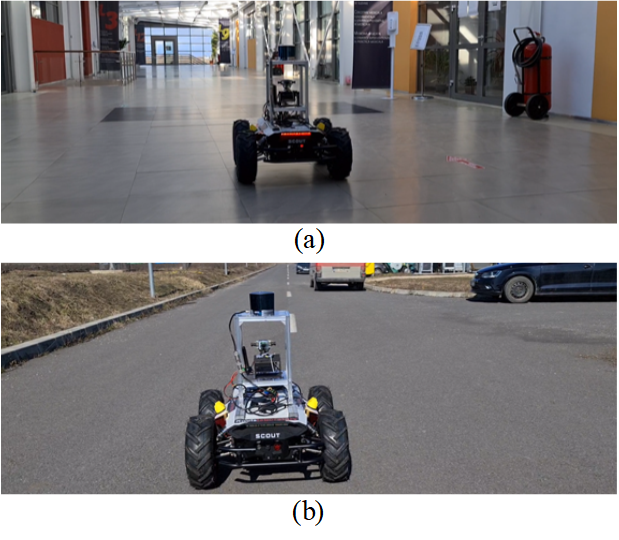}
	    	\vspace{-1em}
		\caption{\textbf{Testing DL-NMPC-SD using RovisLab AMTU in different working conditions.} (a) Indoor environment. (b) Outdoor traffic scene.}
        \label{fig:rovis_lab_amtu_scenes}
	\end{center}
    	\vspace{-1.5em}
\end{figure}

Real-world training data has been collected based on indoor navigation with the RovisLab AMTU system from Fig.~\ref{fig:eb_car}(a). RovisLab AMTU is an AgileX Scout 2.0 platform which acts as 1:4 scaled car, equipped with a $360^{\circ}$ Hesai Pandar 40 Lidar, 4x e-130A cameras providing a $360^{\circ}$ visual perception of the surroundings, a VESC inertial measurement unit and an NVIDIA AGX Xavier board for data processing and control. The robot's test setup parameters are shown in Table~\ref{tab:modelcar_parameters}. Snapshots from the indoor and outdoor testing environment using RovisLab AMTU are shown in Fig.~\ref{fig:rovis_lab_amtu_scenes}.

\begin{table}
	\centering
	\begin{tabular}{ll}
		\hline
		\textbf{Description} & \textbf{Value} \\
		\hline
		Dimensions & $108cm \times 73cm \times 80cm$ \\
		Drive & DC brushless 4 X $150W$ \\
		& 1x Hesai Pandar 40 Lidar \\
		Sensing & 4x e-130A cameras \\
		& 1x VESC inertial measurement unit \\
		Main computing device & 1x Nvidia AGX Xavier \\
		Operating System & Ubuntu Linux 14.04.3 \\
		Middleware SW & Elektrobit ADTF v2.13.2 x64 \\
		\hline
	\end{tabular}
	\caption{RovisLab AMTU parameters.}
	\label{tab:modelcar_parameters}
    	\vspace{-1.5em}
\end{table}

The robot is composed of a differential chassis based on four $150W$ DC brushless motors, enabling it to turn on its yaw axis. Since the robot is not equipped with a radar system, we calculate the OGs solely based on the Lidar data. The four cameras providing a $360^{\circ}$ view of the environment are used for visually inspecting the computed OGs. This is accomplished by projecting the occupancy grid data onto the images using a perspective transformation of the grid. This transformation is based on the fixed extrinsic camera parameters, which define the relationship between the camera coordinates and the robot's body coordinates.

The operating system is an Ubuntu Linux 14.04.3, running Elektrobit's Advanced Data and Time Triggered Framework (ADTF) for developing and running advanced driver assistance systems. The DL-NMPC-SD algorithm is implemented as an ADTF filter, interacting with the robot's sensors, and an actuator filter used for sending steering and velocity commands to the robot. The scene dynamics model is queried in real-time using an inference engine implemented on top of the Open Neural Network eXchange (ONNX) standard.

Approx. $42.000$ samples were acquired by manually driving AMTU. The different dynamics of the environment are given by obstacles and persons interacting with the robot. The interactions involve approaching the robot, overtaking it, or walking alongside it. These movements are mapped to occupancy grids sequences which are further passed as input to the scene dynamics model.

\subsection{Testing Setup using an Autonomous Vehicle}
\label{sec:setup_testing_eb_car}

Real-world testing and data acquisition on public roads has been performed by driving on several types of roads in Erlangen and Tennenlohe, Germany, using the Volkswagen Passat test car shown in Fig.~\ref{fig:eb_car}(b). The vehicle, with its configuration setup detailed in Table~\ref{tab:eb_car_parameters}, is an Elektrobit autonomous test car equipped with a front Continental MFC430 camera, two front and rear Quanergy M8 LiDARs and six front, rear and side Continental ARS430 radars.

\begin{table}
	\centering
	\begin{tabular}{ll}
		\hline
		\textbf{Description} & \textbf{Value} \\
		\hline
		Positioning & 2x Differential GPS \\
		Front and rear LiDARs & 2x Quanergy M8 \\
		Front camera & 1x Continental MFC430 \\
		Front, side and rear radars & 4x Continental ARS430 \\
		Computing system & 64 bit 8-core ARM CPU, 32GB RAM \\
		GPU & Nvidia 512-core Volta GPU \\
		Operating System & Ubuntu Linux 14.04.3 \\
		Middleware SW & Elektrobit ADTF v2.13.2 x64 \\
		\hline
	\end{tabular}
	\caption{Autonomous vehicle parameters.}
	\label{tab:eb_car_parameters}
    	\vspace{-1.5em}
\end{table}

The data acquired from the Quanergy M8 LiDARs and the ARS430 radar is fused in order to calculate the occupancy grids. The practical implementation of the algorithm used to compute the OGs is described in the next Section~\ref{sec:ogs}. As in the case of the tests conducted with RovisLab AMTU, the images acquired from the video cameras are used solely for visual checking.

The computing system of the car uses a 64 bit 8-core ARM CPU, with 32GB RAM, and a 512-core Volta GPU from Nvidia, running Ubuntu Linux and the same ADTF framework setup as the one described in the previous section for the tests with RovisLab AMTU.

\begin{figure}
	\centering
	\begin{center}
		\includegraphics[scale=0.95]{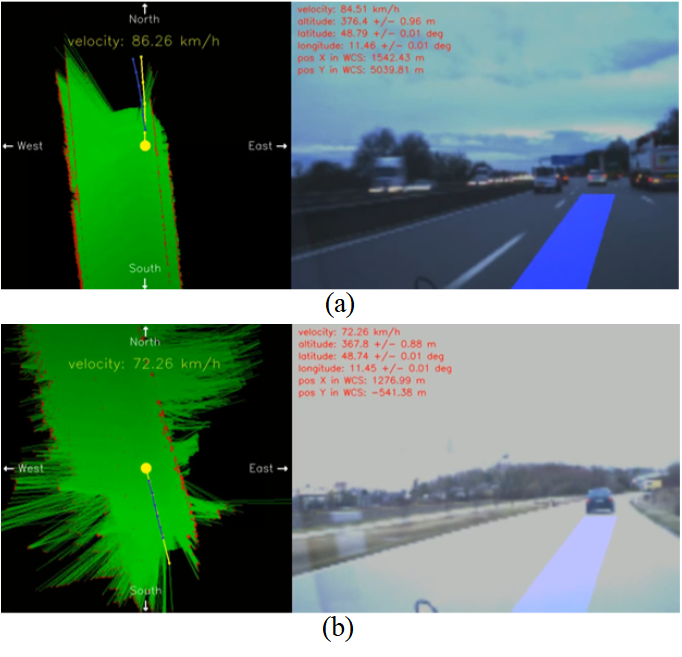}
	    	\vspace{-1em}
		\caption{\textbf{DL-NMPC-SD evaluation using the Elektrobit test car from Fig.~\ref{fig:eb_car}(b), on country roads (a) and highway (b) driving.} The left side of the snapshots illustrate the occupancy grid at time $t$, and the output of the dynamics model (yellow) superimposed on the ground truth trajectory (blue). The right side shows the driving scene imaged through the front camera of the car.}
        \label{fig:eb_car_scenes}
	\end{center}
    	\vspace{-1.5em}
\end{figure}

Approx. $463.000$ samples were acquired in different driving scenarios, including country roads, highways, city driving, T junctions, traffic jams and steep curves. Examples of collected occupancy grids and their corresponding front camera images are shown in Fig.~\ref{fig:occupancy_grids}(c,d), as well as in Fig.~\ref{fig:eb_car_scenes}.

The samples acquired in this test scenario are from areas with high traffic density, such as traffic jams on highways, intersections, roundabouts, bridges and construction zones. They also contain infrequent occurrences, like ambulances, or potentially hazardous traffic situations, such as improper actions by drivers, various driving maneuvers like lane changes, turns, and stops, as well as challenging environmental conditions for an autonomous vehicle.

\section*{Appendix D}
\section*{Data Synchronization and OG Implementation}
\label{sec:ogs}

The sensory data streams are fused into an OG of size $12m \times 12m$, with a $0.1m$ resolution, for the case of RovisLab AMTU, while for the VW test vehicle we use OGs of size $125m \times 125m$, having a $0.25m$ cell resolution. Each data sample is acquired at time intervals in a range between $50ms$ and $90ms$ per cycle.

Considering the world coordinate system as the vehicle's center of mass, the OG always slides along the ego vehicle's driving direction. New radar data is added to the grid using an anchor which describes the position of the grid alongside the driving trajectory. In the sensor's coordinate system, the anchor is located at the top left corner of the OG. The grid's dimension and precision were defined by considering the viewing range of the sensor and a factor used to degrade the occupancy information in grid cells which are not updated by new sensory data.

Due to the variable time intervals used for collecting the data, the world coordinate system is not always positioned in the center of the OG. This is due to the viewing range of the sensor, which is higher than the distance between the ego-car and the last cell in the grid. In the case of driving on roads, the vehicle is able to move approx. $15m$ inside the grid, without re-centering the OG itself. When newly received data falls outside of the grid, the anchor is moved and the OG coordinate system switches position in a forward moving direction. This behavior is problematic when dealing with sequential information inside LSTM networks, since the temporal structure of the input data would not be continuous. In order to overcome this challenge, we have adapted the data acquisition pipeline in order to convert the grid's coordinate system to world coordinates.

\begin{figure}
	\centering
	\begin{center}
		\includegraphics[scale=1.0]{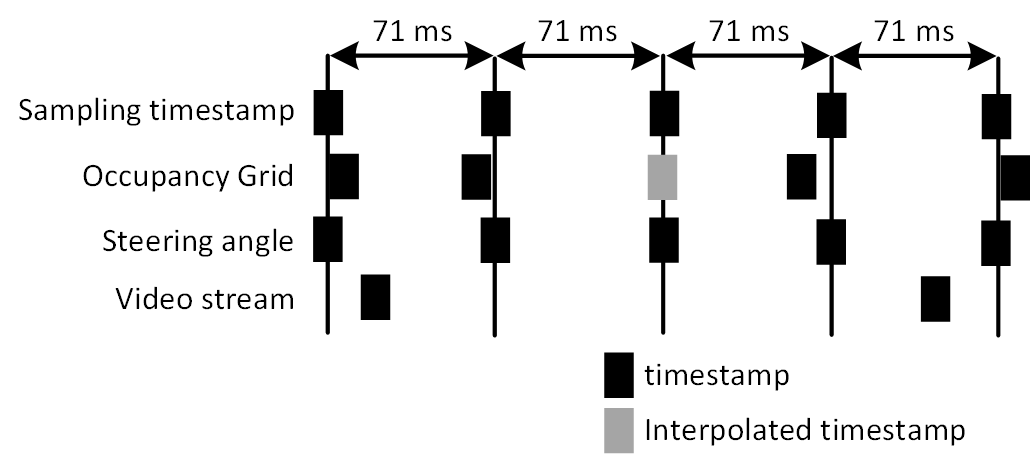}
	    	\vspace{-1em}
		\caption{\textbf{Synchronization of different data sources on the data bus of the test vehicles in Fig.~\ref{fig:eb_car}.} A missing data sample, such as the gray marked OG, is obtained from interpolating the neighboring data elements.}
        \label{fig:timestamp_sync}
	\end{center}
    	\vspace{-1.5em}
\end{figure}

Data acquisition occurs at different timestamps within the car's data recording system. Dynamic information, such as velocity, position of the world coordinate system, or steering angle are saved on the car's data bus with variations of up to $90ms$. Occupancy grids are computed every $50ms$, while image acquisition takes place approximatively every $350ms$. Data storage is performed with a sampling rate of $71ms$. In order to obtain suitable training data, we have implemented the synchronization mechanism illustrated in Fig.~\ref{fig:timestamp_sync}, where missing information is computed via interpolation of neighboring temporal samples.

The camera image is used only for visual validation, so synchronizing it on every sampled timestamp is not required.

\begin{figure*}
	\centering
	\begin{center}
		\includegraphics[scale=0.8]{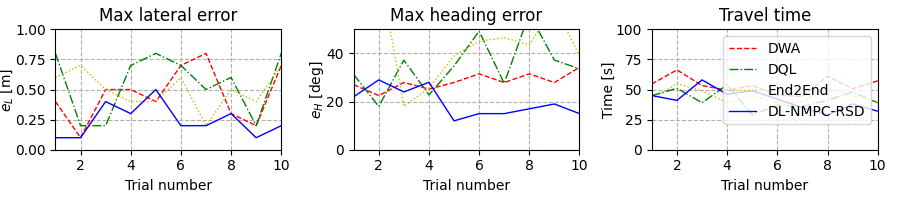}
	    	\vspace{-1.5em}
		\caption{\textbf{Maximum path tracking errors and travel times vs. trial number for experiment \textit{II}}. DL-NMPC-SD results in lower tracking errors and an increased travel time for the 10 experimental trials.}
        \label{fig:max_path_tracking_error_model_car_no_obstacles}
	\end{center}
    	\vspace{-1.5em}
\end{figure*}

\begin{figure*}
	\centering
	\begin{center}
		\includegraphics[scale=0.8]{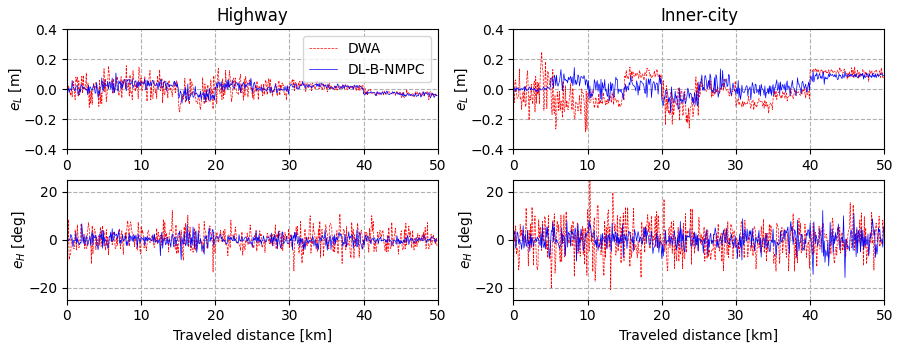}
	    	\vspace{-1em}
		\caption{\textbf{Path tracking errors vs. travel distance for DWA and DL-NMPC-SD experiments performed with the test car from Fig.~\ref{fig:eb_car}(b)}. Due to the complex nature of the driving environment, the measured errors are higher for the case of inner-city driving, in comparison to highway driving.}
        \label{fig:path_tracking_real_car}
	\end{center}
    	\vspace{0.5em}
\end{figure*}

\section*{Appendix E}
\section*{Implementation of Competing Algorithms}
\label{sec:competing_algorithms}

The three competing algorithms used to evaluate DL-NMPC-SD has have been implemented as follows.

\subsubsection{DWA}
Dynamic Window Approach~\cite{Fox_Dynamic_Window_Approach_1997} is an online collision avoidance strategy for mobile robots, which uses robot dynamics and constraints imposed on the robot's velocities and accelerations to calculate a collision free path in the 2D plane. We have implemented DWA based on the Robot Operating System's (ROS) DWA local planner. DWA takes as input the distances from the ego-vehicle to the obstacles present in the scene, calculated as perception rays.

\subsubsection{End2End Learning}
In the case of End2End learning~\cite{10614862_end2end}, we have mapped sequences of OGs directly to the discrete driving commands of the vehicle (turn left, turn right, accelerate, decelerate), in a behavioral cloning fashion. Although continuous control is possible, for example based on policy gradient estimation~\cite{Lillicrap2016ContinuousCW}, the most common strategies for end2end and DRL in autonomous driving are based on discrete control~\cite{Jaritz2018}.

The network topology is based on~\cite{10614862_end2end}, where the final soft-max layer predicts the four discrete steering commands as a classification output. The steering commands are executed with an incremental value of $0.01$, dependent on the End2End system's output (e.g. if the prediction output is "turn left", then the steering angle of the car is incremented with a $0.01$ value).

\subsubsection{DQN}
The benchmarking against DQL was performed with respect to the Deep Q-Learning (DQN) algorithm described in~\cite{mnih2015humanlevel, 9904958_DRL_Survey}. Due to the fact that in a reinforcement learning framework the agent gets trained through its interaction with its working environment, we have trained the DQN system based on GridSim simulations only. The DQN agent was trained with the following hyperparameters: 

\begin{itemize}
	\item memory size: $10^6$;
	\item discount rate $\gamma$: $0.95$;
	\item initial greedy factor $\epsilon_0 = 1.0$ ;
	\item decay value: $\epsilon_{decay} = 0.99$;
	\item minimum greedy factor: $\epsilon_{min} = 0.05$;
	\item learning rate: $25 \times 10^{-7}$;
	\item L2 regularization factor: $\lambda_2 = 10^{-4}$.
\end{itemize}

The input data is a $4$-dimensional tensor represented by a sequence of $4$ images from the GridSim simulator, with a sampling difference of $5$ frames. The frame data was downscaled to $80px \times 80px$ for speed optimization. The linear activation of the last layer of the DQN network produces an integer mapped to a decision space of eight actions: accelerate, turn left, turn right, no action, brake, accelerate + turn left, accelerate + turn right and reverse/negative acceleration.

We have constructed a reward policy, where the total reward is normalized to the $[-1, 1]$ interval. The reward function is a weighted sum composed of the distance traveled by the agent at time step $t$, the current velocity of the vehicle and the sensor policy. The minimum value of the artificial sensor is then compared to several thresholds that produce an acceptance score, where $1.0$ is the ideal score and $-1.0$ is the worst. To avoid little to no movement after taking an action, we have implemented a sanity check which takes into consideration the distance traveled and current speed. As in the case of End2End learning, the steering commands are executed based on the $0.01$ increment value.

\section*{Appendix F}
\section*{Path Tracking Errors}
\label{sec:competing_algorithms}

\subsection{Indoor and Outdoor Testing Results}

Fig.~\ref{fig:max_path_tracking_error_model_car_no_obstacles} illustrates the maximum tracking errors and travel time vs. trial number are shown in Appendix~F. The results have been obtained from $5$ trials, with no obstacles on the ground truth path, as well as from $5$ trials containing static and dynamic obstacles. $6$ of the trials have been performed indoors, while $4$ of them have been conducted outdoors, using the environments detailed in Appendix~C.

\subsection{On-Road Testing Results}

The difference in path tracking errors (Fig.~\ref{fig:path_tracking_real_car}), between highway and inner-city driving, comes from the unstructured nature of inner-city environments, since the scenes' variance is lower in highway driving, when compared to the inner-city environment. In practice, we do not expect DL-NMPC-SD's tracking errors to go to zero, since the algorithm balances in an optimal control fashion tracking errors and control inputs subject to constraints.

\end{document}